\pdfoutput=1
\documentclass[11pt]{article}

\usepackage[preprint]{acl}

\usepackage{times}
\usepackage{latexsym}

\usepackage[T1]{fontenc}

\usepackage[utf8]{inputenc}

\usepackage{microtype}

\usepackage{inconsolata}

\usepackage{graphicx}
\usepackage{amsmath}
\usepackage{algorithm}
\usepackage{algpseudocode}
\usepackage{booktabs}

\usepackage{times}
\usepackage{booktabs}
\usepackage{enumitem}
\usepackage{multirow}
\usepackage{soul}
\usepackage{pgfplots}
\usepackage{hyperref}
\usepackage{url}
\usepackage{subcaption}
\usepackage[most]{tcolorbox}
\usepackage{multicol}

\usepackage{physics}
\usepackage{verbatim}
\usepackage{amssymb}
\usepackage{pifont}
\usepackage{ulem}
\usepackage{mathrsfs}
\usepackage{xspace}
\usepackage{arydshln}
\usepackage{float}

\useunder{\uline}{\ul}{}
\usepackage{colortbl}
\PassOptionsToPackage{prologue,dvipsnames}{xcolor}
\usepackage[dvipsnames]{xcolor}
\usepackage{pgfmath}

\newcommand{\mkj}{{\usefont{T1}{ppl}{m}{n}MKJ}\xspace}

\definecolor{codegreen}{rgb}{0,0.6,0}
\definecolor{codegray}{rgb}{0.5,0.5,0.5}
\definecolor{codepurple}{rgb}{0.58,0,0.82}
\definecolor{backcolour}{rgb}{0.95,0.95,0.92}
\usepackage{listings}

\lstdefinestyle{mystyle}{
  commentstyle=\color{codegreen},
  keywordstyle=\color{magenta},
  numberstyle=\tiny\color{codegray},
  stringstyle=\color{codepurple},
  basicstyle=\ttfamily\footnotesize,
  breakatwhitespace=false,
  breaklines=true,
  captionpos=b,
  keepspaces=false,
  showspaces=false,
  showstringspaces=false,
  showtabs=false,
  tabsize=2
}

\usepackage{microtype}
\usepackage{cleveref}

\newcommand{\gpts}{\textsc{GPT-Series}\xspace}
\newcommand{\gptthree}{\textsc{GPT-3.5-Turbo}\xspace}
\newcommand{\gptfour}{\textsc{GPT-4o-mini}\xspace}
\newcommand{\gptfouro}{\textsc{GPT-4o}\xspace}

\newcommand{\claudes}{\textsc{Claude-3-Series}\xspace}
\newcommand{\claudea}{\textsc{Claude-3-haiku}\xspace}
\newcommand{\claudeb}{\textsc{Claude-3-sonnet}\xspace}

\newcommand{\llamas}{\textsc{Llama-3-Series}\xspace}
\newcommand{\llamaa}{\textsc{Llama-3-8B}\xspace}
\newcommand{\llamab}{\textsc{Llama-3.1-8B}\xspace}
\newcommand{\llamac}{\textsc{Llama-3.2-1B}\xspace}
\newcommand{\llamad}{\textsc{Llama-3.2-3B}\xspace}

\newcommand{\ministral}{\textsc{Ministral-8B}\xspace}

\newcommand{\qwens}{\textsc{Qwen2.5-Series}\xspace}
\newcommand{\qwena}{\textsc{Qwen2.5-0.5B}\xspace}
\newcommand{\qwenb}{\textsc{Qwen2.5-1.5B}\xspace}
\newcommand{\qwenc}{\textsc{Qwen2.5-3B}\xspace}

\newcommand{\phis}{\textsc{Phi-3-Series}\xspace}
\newcommand{\phia}{\textsc{Phi-3-mini-4k}\xspace}
\newcommand{\phib}{\textsc{Phi-3-mini-128k}\xspace}
\newcommand{\phic}{\textsc{Phi-3.5-mini}\xspace}

\newcommand{\meditron}{\textsc{Meditron-7B}\xspace}
\newcommand{\mellama}{\textsc{MeLlama-13B}\xspace}

\newcommand{\diff}[2]{
  \pgfmathsetmacro{\difference}{#1 - #2}%
  \pgfmathsetmacro{\absdiff}{abs(\difference)}%
  \ifdim \difference pt>0pt
    \textcolor{green}{\scriptsize ($\uparrow$ \absdiff)}%
  \else
    \ifdim \difference pt<0pt
      \textcolor{red}{\scriptsize ($\downarrow$ \absdiff)}%
    \else
      (0.00)%
    \fi
  \fi
}

\title{Fact or Guesswork? Evaluating Large Language Models' Medical Knowledge with Structured One-Hop Judgments}

\author{
Jiaxi Li$^1$ \quad Yiwei Wang$^2$ \quad Kai Zhang$^3$ \quad Yujun Cai$^4$ \\
\textbf{Bryan Hooi$^5$ \quad Nanyun Peng$^6$ \quad Kai-Wei Chang$^6$ \quad Jin Lu$^1$} \\ 
$^1$University of Georgia \quad 
$^2$University of California, Merced \\
$^3$Lehigh University \quad
$^4$The University of Queensland \\
$^5$National University of Singapore \quad
$^6$University of California, Los Angeles \\
\href{https://plusnli.github.io/med-knowledge-judgment.github.io/}{\textcolor{magenta}{\texttt{plusnli.github.io/med-knowledge-judgment.github.io}}}
}

\pgfplotsset{compat=1.18}

\begin{document}

\maketitle

\begin{abstract}
Large language models (LLMs) have been widely adopted in various downstream task domains.
However, their abilities to directly recall and apply factual medical knowledge remains under-explored. Most existing medical QA benchmarks assess complex reasoning or multi-hop inference, making it difficult to isolate LLMs' inherent medical knowledge from their reasoning capabilities. 
Given the high-stakes nature of medical applications, where incorrect information can have critical consequences, it is essential to evaluate the factuality of LLMs to retain medical knowledge.

To address this challenge, we introduce the \textbf{Medical Knowledge Judgment Dataset (\mkj)}, a dataset derived from the Unified Medical Language System (UMLS), a comprehensive repository of standardized biomedical vocabularies and knowledge graphs. Through a binary classification framework, \mkj evaluates LLMs' grasp of fundamental medical facts by having them assess the validity of concise, one-hop statements, enabling direct measurement of their knowledge retention capabilities.

Our experiments reveal that LLMs have difficulty accurately recalling medical facts, with performances varying substantially across semantic types and showing notable weakness in uncommon medical conditions.
Furthermore, LLMs show poor calibration, often being overconfident in incorrect answers. 
To mitigate these issues, we explore retrieval-augmented generation, demonstrating its effectiveness in improving factual accuracy and reducing uncertainty in medical decision-making.

\end{abstract}

\section{Introduction}
Large language models (LLMs) such as GPT-4~\cite{openai2023gpt4} and Llama3~\cite{dubey2024llama} have demonstrated remarkable generative capabilities, achieving state-of-the-art performance in various NLP tasks.
However, their abilities to accurately recall and apply domain-specific knowledge remain a major challenge, particularly in high-stakes fields such as medicine and healthcare~\cite{li2022neural, yang2023large, liu2023evaluating, yan2023multimodal}. LLMs are prone to hallucinations - generating plausible but incorrect content~\cite{bang2023multitask, zhang2023siren, liu2025mitigating}, which raises concerns about their factuality~\cite{li-etal-2024-dawn, latif2024systematic}. 
These domain-specific knowledge gaps often lead to contradictions, misinformation, and overconfident yet incorrect predictions, making them unsuitable for direct deployment in clinical decision-making. 

\begin{figure}[t!]
\centering
\includegraphics[width=0.48\textwidth]{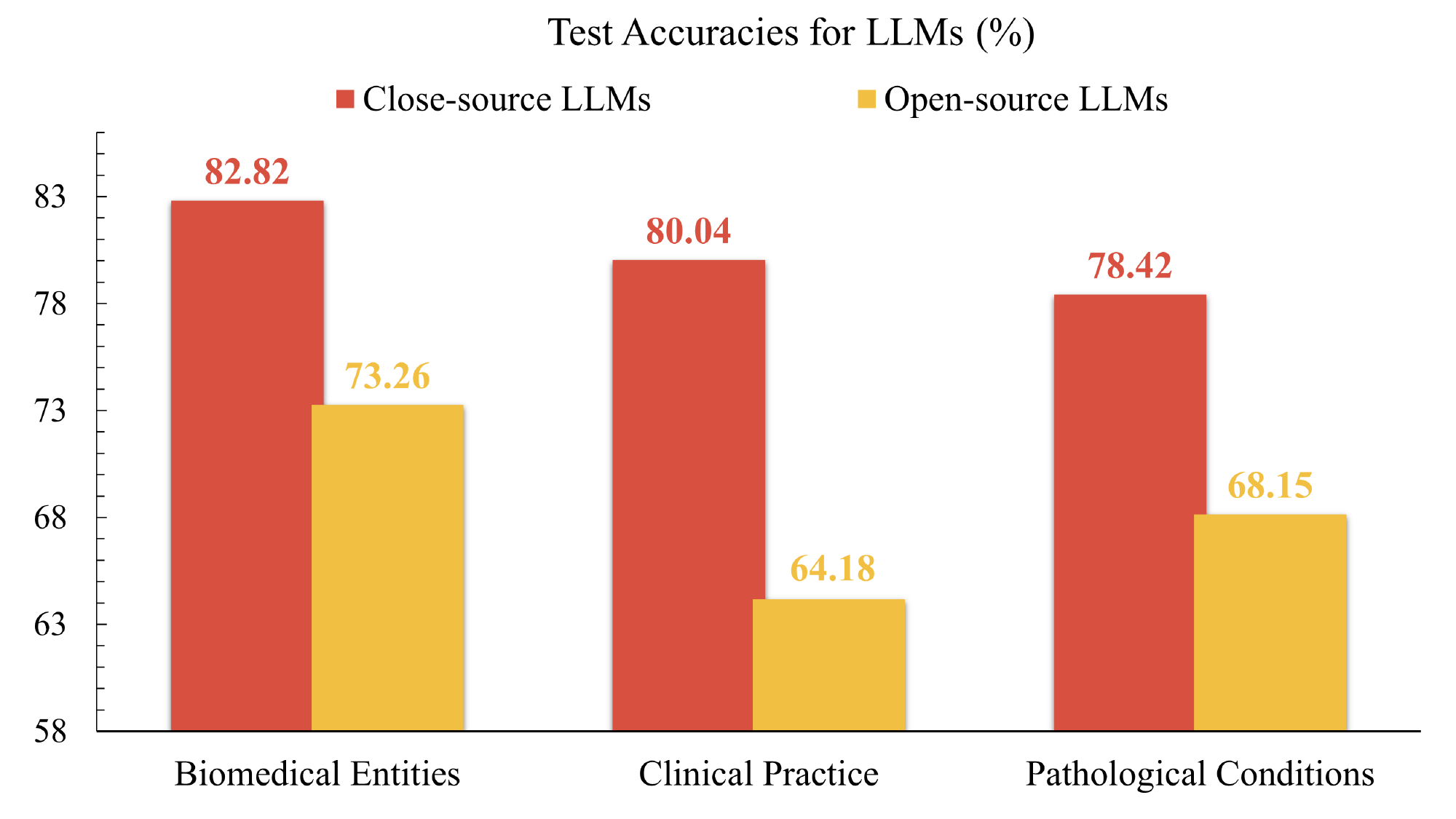} 

\vspace{4pt}

\scriptsize
\begin{tabular}{p{0.45\textwidth}}
    \textbf{Example cases on which \gptfour fails} \\
    
    \toprule
    \textbf{Semantic Type:} \textbf{\textit{Neoplastic Process}} \\
    \textbf{Judgment:} Gastrointestinal Tract is primary anatomic site of disease Extrahepatic Bile Duct Tubular Adenoma. \\
    \textbf{Label:} True \\
    \textbf{\gptfour:} False \\

    \midrule
    \textbf{Semantic Type:} \textbf{\textit{Clinical Drug}} \\
    \textbf{Judgment:} The multivitamin with minerals Calcium and Magnesium oral tablet contains the ingredient Lipitor. \\
    \textbf{Label:} False \\
    \textbf{\gptfour:} True \\
    
    \midrule
    \textbf{Semantic Type:} \textbf{\textit{Hormone}} \\
    \textbf{Judgment:} The codeine hydrochloride is a contraindicated drug for the Deep Vein Thrombosis. \\
    \textbf{Label:} False \\
    \textbf{\gptfour:} True \\
    
    
    \bottomrule
\end{tabular}
\caption{Accuracy for close-source and open-source LLMs evaluated on the \mkj dataset across three categories. Examples are provided that \gptfour fails to answer correctly.} 
\label{fig:overview}
\vspace{-1.8em}

\end{figure}

To evaluate \textit{``how knowledgeable are LLMs in medicine and healthcare''}, existing benchmarks typically frame the task as a question-answering (QA) challenge, often involving clinical questions that require multi-step reasoning, indirect relationships, or external retrieval~\cite{pal2022medmcqa, malaviya-etal-2024-expertqa, zhou2025automating}. 
However, evaluating LLMs' core medical knowledge requires a direct and controlled framework, one that can systematically quantify what LLMs ``know'' without the confounding effects of reasoning or retrieval augmentation.
Fundamental knowledge evaluation is not merely an auxiliary task. It is a prerequisite for trustworthy and effective medical foundation models~\cite{zhang2024generalist, moor2023foundation}. By ensuring that LLMs possess a solid factual foundation, we pave the way for more reliable reasoning, clinical applications, and ultimately, safer deployment in real-world healthcare settings.

To bridge this gap, we introduce the Medical Knowledge Judgment Dataset (\mkj), which is designed to systematically evaluate LLMs' inherent factual medical knowledge through one-hop binary judgment task. 
We construct \mkj through a systematic process of extracting knowledge triplets from Unified Medical Language System (UMLS) and transforming them into carefully templated judgment questions. 
We focus on one-hop relation and remove triplets with multi-hop nodes or multiple relationships, which ensures single and indisputable answer~\cite{wei2024measuring, sun2024head}. Some examples are provided in Figure~\ref{fig:overview}.

The UMLS serves as the ideal foundation for our dataset due to its unparalleled comprehensiveness and reliability in the medical domain. 
First, UMLS is a rigorously curated and widely trusted biomedical resource, integrating over 3.8 million concepts and 78 million relationships from authoritative medical terminologies, which ensures the knowledge assessed in our dataset is clinically validated and standardized. 
Second, its knowledge graph (KG) structure provides an explicit, structured representation of medical knowledge, enabling precise fact-based evaluation while minimizing ambiguity and inconsistencies often found in alternative sources~\citep{abacha2017overview, malaviya-etal-2024-expertqa}. 
By leveraging UMLS as a gold-standard knowledge base, we ensure high factual reliability, broad medical coverage, and systematic evaluation of LLMs' medical factuality. The generated questions are then grouped into three progressive categories: Biomedical Entities (foundational concepts), Pathological Conditions (phenomena), and Clinical Practice (applications) with detailed information in Appendix~\ref{app:data_detail}.


To comprehensively examine LLMs' capabilities in medical knowledge retention, we focus on four specific research questions (RQs):

\begin{itemize}[left=0pt, topsep=0pt, itemsep=0pt, partopsep=0pt,parsep=0pt]
    \item \textbf{RQ1:} To what extent can LLMs accurately perform medical judgments?
    \item \textbf{RQ2:} How well are LLMs calibrated in medical and healthcare contexts?
    \item \textbf{RQ3:} What are the underlying reasons behind LLMs' failure to retain certain critical medical knowledge?
    \item \textbf{RQ4:} What strategies can enhance the response accuracy of LLMs?
\end{itemize}

Through comprehensive analysis on \mkj, we find that LLMs, especially open-source LLMs, still struggle with basic factual medical knowledge retention as illustrated in Figure~\ref{fig:overview}. 
Our investigation in later sections reveals critical challenges: poor calibration with frequent overconfidence in incorrect predictions as shown in Figure~\ref{fig:calibration_curves}, and notably degraded performance when handling rare medical conditions as shown in Figure~\ref{fig:acc_and_freq}. 
To address these limitations, we implement retrieval-augmented generation, which substantially improves factual accuracy as displayed in Table~\ref{tab:sparse_dense_rag}.

Our work's contributions extend beyond the introduction of the \mkj dataset. 
The systematic evaluation and analysis provide comprehensive insights into LLMs' handling of medical knowledge, illuminating challenges and opportunities that can inform future medical applications.

\section{Related Work}

\begin{figure*}[!t]
    \centering
    \vspace{-1.6em}
    \resizebox{\textwidth}{!}{
        \includegraphics[width=\textwidth]{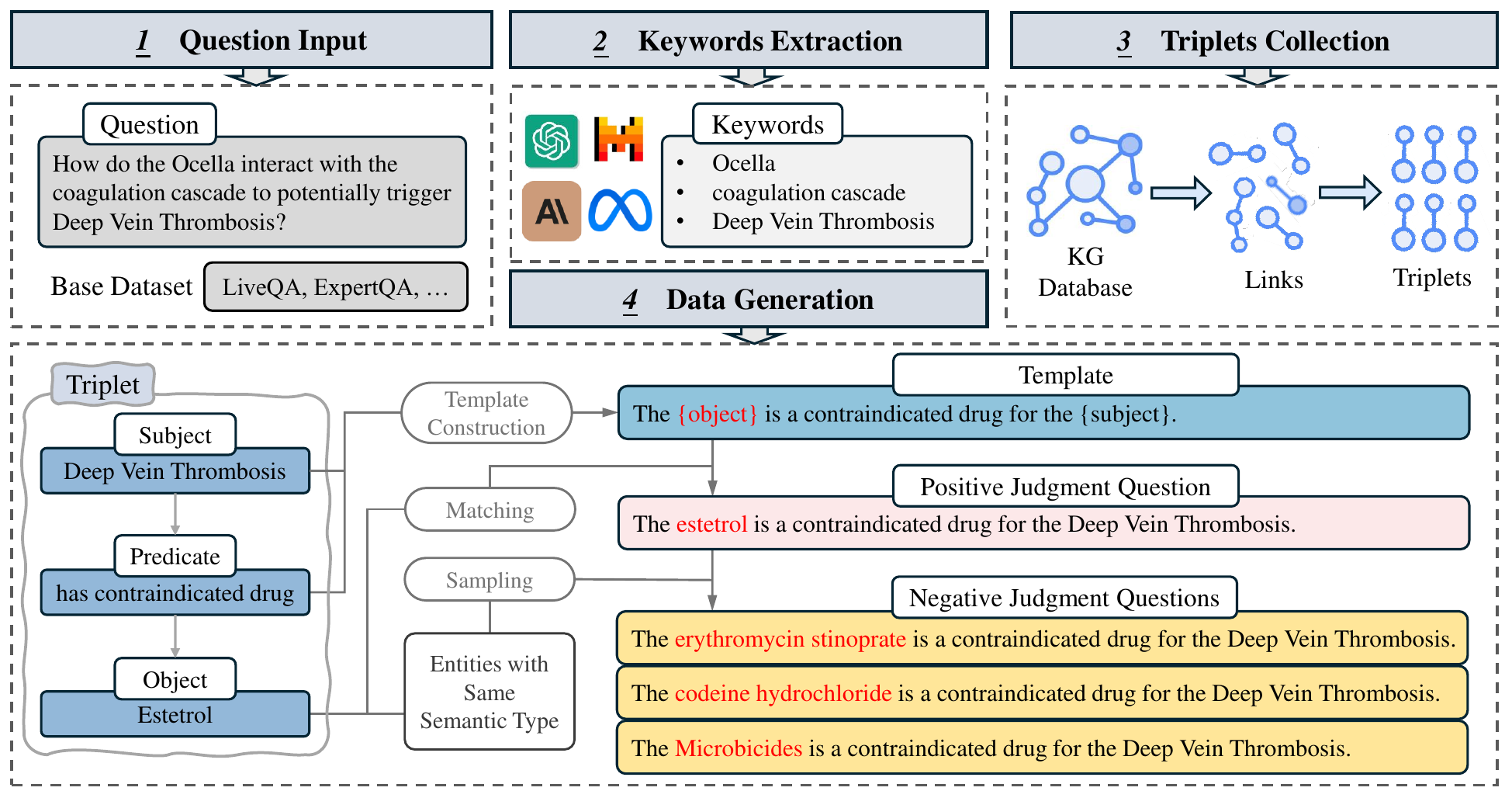}
    }
    \caption{Pipeline for the \mkj dataset construction.}
    \label{fig:pipeline}
    \vspace{-0.6em}
\end{figure*}

\paragraph{Factuality Benchmarks.}
Evaluating LLM factuality presents a non-trivial challenge and various benchmarks are proposed in the general domain~\cite{joshi2017triviaqa, trivedi2017lc, dubey2019lc, lin2022truthfulqa, sun-etal-2024-head, kim-etal-2023-factkg}.
Recently, the SimpleQA series of works~\cite{wei2024measuring, cheng2025simplevqa, gu2025chinesesimplevqa, he2024chinese} facilitate factuality evaluation by constraining the scope to short, fact-seeking questions with single answers, making factuality assessment more tractable compared to previous long and open-ended model outputs.

Despite the progress, the factuality evaluation in medicine and healthcare for LLMs nowadays is still a lack to some extent. 
There are some prior work exploring direct probing of medical knowledge in language models~\cite{meng-etal-2022-rewire, zhou2024reliable}. However, they either rely on cloze-style queries that may inadvertently assess reasoning rather than pure knowledge, or employ LLM-based generation methods that could introduce noise and bias. 
In contrast, our \mkj dataset employs binary classification questions with a systematic construction pipeline based on entity substitution, providing a more direct and unambiguous evaluation of medical knowledge in LLMs.

\paragraph{LLMs in Medicine.}
To apply LLMs on different tasks in medical and healthcare contexts, some approaches are adopted such as prompt engineering~\cite{shi2024ehragent, chen2024cod, singhal2025toward} and fine-tuning LLMs with domain-specific data~\cite{xu2023baize, xie2024me, chen2023meditron70b, shi2024mgh, xiong2023doctorglm, zhao2024helene}.
There are also some work integrating medical knowledge graphs to assist LLMs on tasks such as medical question answering~\cite{yang2024kg, yasunaga2022deep} and diagnosis prediction~\cite{afshar2024role, gao2023leveraging}.
While existing efforts have focused on enhancing LLMs' performances across various tasks, fundamental assessments of their capabilities to internalize and apply medical knowledge remain relatively limited. 
Our proposed \mkj dataset bridges this gap by constructing questions covering various factual medical knowledge types, and answers in the easily verifiable short-form format.

\section{Dataset Construction}
\label{sec:dataset}

\subsection{Overview}
The objective of \mkj dataset is to provide a structured and controlled environment for evaluating LLMs on fundamental, one-hop medical knowledge. We use the following three-step construction pipeline, which yields $3,000$ rigorously curated and standardized questions:

\begin{enumerate}[left=0pt,topsep=0pt,partopsep=0pt,parsep=0pt,itemsep=0pt]
    \item \textbf{Triplet Extraction.} We systematically retrieve medical knowledge triplets $(s, p, o)$ from the UMLS, then apply filtering steps to remove substandard or repetitive entries.
    \item \textbf{Template Design.} We assess the evaluation as a binary classification task, which involves constructing predicate-specific templates that map each triplet into a well-defined judgment statement (either "True" or "False").
    \item \textbf{Entity Substitution.} We populate the templates with entities of the same semantic types to generate both positive and negative samples for creating balanced binary judgments.
\end{enumerate}

    
    

\subsection{Triplets collection}
\label{sec:data_collect}
Let $\mathcal{G}=\{\mathcal{V}, \mathcal{E}\}$ represent the medical knowledge graph derived from UMLS, where $\mathcal{V}$ is the set of vertices (i.e., medical concepts or entities), and $\mathcal{E}$ is the set of predicates (relations) among them.

Concretely, UMLS contains over 3.8 million concepts $|\mathcal{V}|$ and more than 78 million relations $|\mathcal{E}|$. Since directly using all possible triples is infeasible, we filter and refine relavent samples as follows:

1. \textbf{Preliminary Question Set.} We begin with a set of multi-hop medical QA benchmarks $\mathcal{Q}_{ori}$, such LiveQA~\cite{abacha2017overview} and ExpertQA~\cite{malaviya-etal-2024-expertqa}. Each question $Q\in \mathcal{Q}_{ori}$ is a natural-language query. It provides a diverse set of real-world medical problems and serve as a basis for identifying the most relevant and valuable UMLS knowledge components.

2. \textbf{Keyword Extraction.} For each $Q$, we prompt an LLM (e.g. \gptfour~\cite{openai2023gpt4} ) to recognize specialized medical named entities. Let 
\begin{equation*}
\mathcal{K}(Q)=\{k_1,\dots,k_n\}
\end{equation*}
be the extracted set of medical keywords from $Q$.

3. \textbf{Graph Matching.} For each keyword $k\in\mathcal{K}(Q)$, we query the UMLS API\footnote{https://documentation.uts.nlm.nih.gov/rest/home.html} to acquire associated links 
\begin{equation*}
\ell=(v_1, e, v_2),
\end{equation*}
where $v_1, v_2\in \mathcal{V}$ and $e\in \mathcal{E}$. Denote the multiset of all such links retrieved across all $Q \in \mathcal{Q}_{ori}$ by $\mathcal{L}$.

4. \textbf{Deduplication \& Filtering.} We deduplicate entries and incorrect artifacts from $\mathcal{L}$. This process yields a high-quality subset:
\begin{equation*}
\mathcal{T}=\{ (s,p,o) \mid s,o \in \mathcal{V}, p\in \mathcal{E} \},
\end{equation*}
where each element $(s,p,o)$ corresponds to a verified medical subject-predicate-object triplet.

5. \textbf{Knowledge Base Construction.} Finally, each triplet $(s,p,o)$ is converted into a short  ``fact'' in natural language (e.g., ``s is a contraindicated drug for o''), which forms a knowledge base $\mathcal{D}$. This textual representation serves as the source for subsequent experiments in later sections.

\subsection{Template construction}
\label{sec:data_temp}
To facilitate direct assessment of LLMs' medical knowledge, we frame our evaluation as a binary classification task, converting extracted triplets $\mathcal{T}$ into binary judgment questions.

Central to this transformation is our template-based approach, where we develop a specific template for each predicate type $p$ in the triplet $(s, p, o)$, which enables us to consistently frame the medical knowledge contained in each triplet as a well-structured binary judgment question, and facilitates standardized precise evaluation.

Specifically, we identify about 200 unique predicates among the collected triplets $\mathcal{T}$. Then we leverage LLMs to generate and human experts to check the templates tailored to the unique predicates present in the triplets $\mathcal{T}$.

\subsection{Question generation with templates}
\label{sec:data_sub}
Having predicate-specific templates, we proceed to generate binary judgment questions by applying these templates to the UMLS triplets $\mathcal{T}$. 
This process involves substituting the entities from each triplet into the corresponding placeholders in the template to form complete and meaningful judgments. The judgments involve positive judgments that their statements hold, as well as negative judgments that their statements do not hold.

For positive judgment samples, they are generated by simply replacing the placeholders in the templates with the original entities present in the triplet. 

However, for negative judgment samples, we need to construct statements that do not hold. In practice, we substitute the object placeholder with other entities that do not satisfy the given relationship with the subject in the triplet. It is essential to ensure that the substituted entities belong to the same semantic type as the original object to maintain logical coherence.
To achieve this, we collect $N_i$ entities within the same semantic type $i$ as the object in the original triplet. We then randomly shuffle this list and select $k$ candidate entities (where $k \ll N_i$) and fill in the placeholder to create $k$ negative ones (here we set $k=3$). 

While negative samples are generated by randomly sampling objects from the same semantic type, there remains a nonzero yet small probability that the resulting statement $(s,p,o_{\text{neg}})$ might actually reflect a valid relation. 
To solve this issue, we perform post-hoc checks against our curated knowledge base $\mathcal{D}$. In the rare case where $(s,p,o_{\text{neg}})$ is present in $\mathcal{D}$, we exclude that instance and resample. In practice, we find this probability to be negligible (below 1\%), which is partially due to the large amount of entities within the same semantic type as shown in Appendix~\ref{app:data_detail}, ensuring that nearly all sampled negative statements are indeed invalid relationships.

The resulting set of binary judgment questions forms the \mkj{} dataset, providing resource for evaluating the medical knowledge within LLMs. Using templates to guide question generation, we enable robust and interpretable assessments across a wide range of aspects of medical knowledge. More information about the dataset can be fonud in Appendix~\ref{app:data_detail}.

\section{Experiments and Analysis}
\label{sec:exp}

\begin{figure*}[!ht]
    \vspace{-0.8em}
    \centering
    \resizebox{0.85\textwidth}{!}{
        \includegraphics[width=\textwidth]{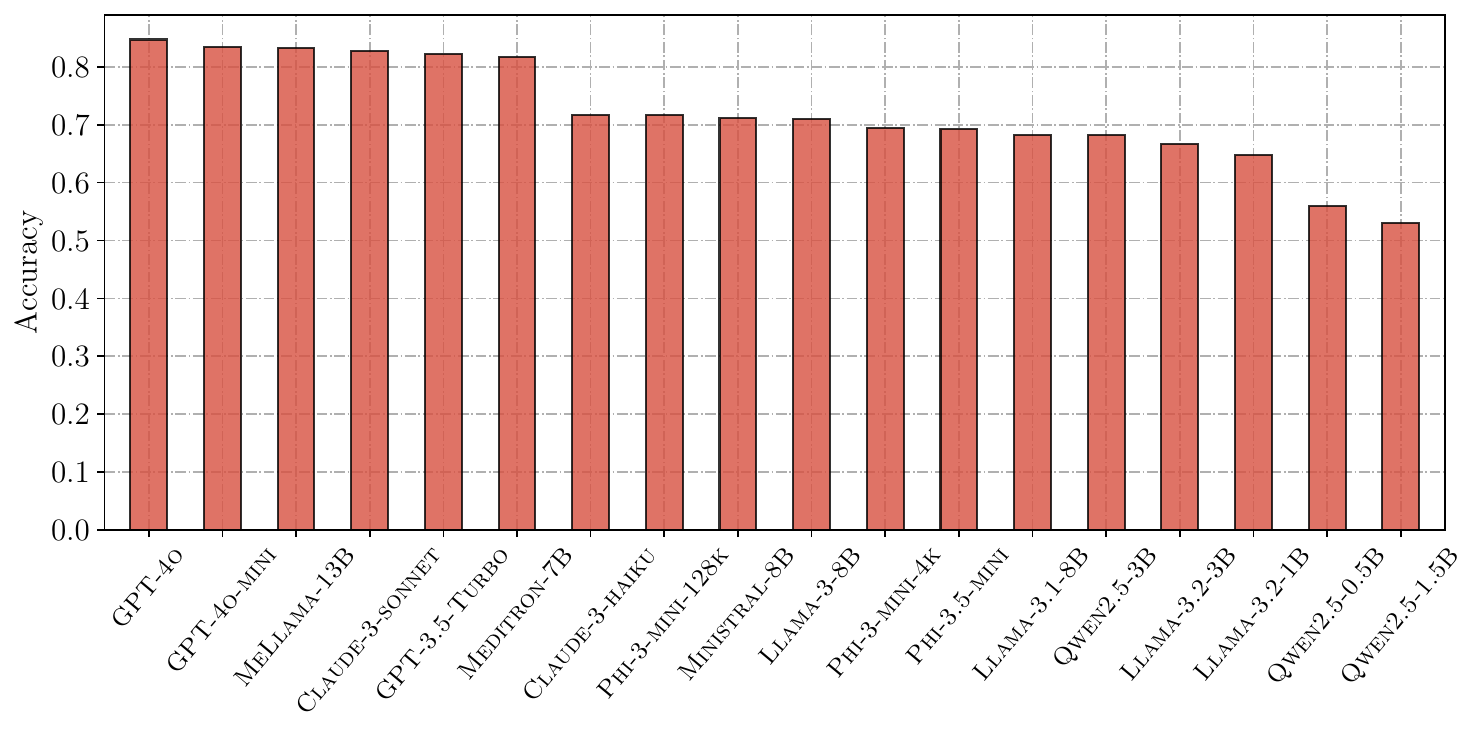}
    }
    \caption{Accuracies of LLMs on the \mkj dataset with zero-shot prompting.}
    \label{fig:zo_performaces}
    \vspace{-0.4em}
\end{figure*}

\subsection{Experiment setup}
We evaluate representative open-source and closed-source LLMs of various sizes and architectures, with an intention on LLMs with smaller sizes. Closed-source models include \gptthree, \gptfour, \gptfouro~\cite{openai2023gpt4}, \claudea and \claudeb~\cite{claude2024}. 
Open-source LLMs involve Llama-3 series (including \llamaa, \llamab, \llamac, \llamad)~\cite{dubey2024llama}, \ministral~\cite{ministral2024}, Qwen2.5 series (\qwena, \qwenb, \qwenc)~\cite{qwen2.5}, Phi-3 series (including \phia, \phib, \phic)~\cite{abdin2024phi}, as well as medical LLMs \meditron~\cite{chen2023meditron70b} and \mellama~\cite{xie2024me}.

More information can be found in Appendix~\ref{app:model_details}.
We utilize zero-shot prompting technique to LLMs for their responses and employ the most deterministic setting (i.e., \texttt{temperature=0} or \texttt{top\_k=1}).
Detailed prompts can be found in Appendix~\ref{app:prompt}.

\subsection{Evaluation}
Each datapoint in the \mkj{} dataset contains 1 positive judgments and $k$ negative judgments (we set $k$=3). Therefore, we have three categories of evaluation metrics, positive accuracy $\text{Acc}_\text{pos}$ (accuracy on the positive judgment questions), negative accuracy $\text{Acc}_\text{neg}$ (accuracy on the negative judgment questions), and accuracy $\text{Acc}$ (regular accuracy over all samples in $\text{Acc}_\text{pos}$ and $\text{Acc}_\text{neg}$ combined), which can be formally defined as:
\begin{equation*}
\text{Acc} = \frac{1}{k+1} \text{Acc}_\text{pos} + \frac{k}{k+1} \text{Acc}_\text{neg}.
\end{equation*}


\subsection{RQ1: To what extent can LLMs accurately perform medical judgments?}
\label{sec:rq1}
Figure~\ref{fig:zo_performaces} shows the overall accuracies Acc of LLMs with zero-shot prompting, from which we can observe that \gptfouro and \gptfour achieve the best results, and medical LLMs \mellama and \meditron also display strong performances. 
However, general-domain open-source LLMs fall behind and their performances generally decreases as model sizes becomes smaller.

\begin{table}[!h]
    \centering
    \resizebox{0.48\textwidth}{!}{
    \begin{tabular}{lcccc}
        \toprule
        Model & Acc & $\text{Acc}_\text{pos}$ & $\text{Acc}_\text{neg}$ & $\text{Acc}_\text{gap}$ \\
        
        \midrule
        \gpts    & 0.83 & 0.69 & 0.88 & -0.19 \\
        \claudes & 0.77 & 0.78 & 0.77 &  0.01 \\
        \ministral & 0.71 & 0.48 & 0.79 & -0.31 \\
        \llamas  & 0.68 & 0.39 & 0.77 & -0.38 \\
        \qwens   & 0.59 & 0.46 & 0.63 & -0.17 \\
        \phis    & 0.70 & 0.63 & 0.73 & -0.10 \\
        
        \meditron & 0.82 & 0.74 & 0.84 & -0.10 \\
        \mellama  & 0.83 & 0.79 & 0.85 & -0.06 \\
        
        \bottomrule
    \end{tabular}
    }
    \caption{Averaged results of accuracy metrics for LLMs and their corresponding gaps $\text{Acc}_\text{gap}$, measured as the difference between positive ($\text{Acc}_\text{pos}$) and negative ($\text{Acc}_\text{neg}$) accuracies.}
    \label{tab:acc_gap}
\end{table}



The results of positive accuracies $\text{Acc}_\text{pos}$ and negative accuracies $\text{Acc}_\text{neg}$ are shown in Table~\ref{tab:acc_gap}. 
We also calculate the accuracy gap between $\text{Acc}_\text{pos}$ and $\text{Acc}_\text{neg}$, which is denoted as $\text{Acc}_\text{gap} = \text{Acc}_\text{pos} - \text{Acc}_\text{neg}$. 
For LLM families with more than one models tested in our experiment, we report the averaged results (round to two decimal places) for convenience and denote these aggregated values with the suffix \textsc{-Series} (e.g., \llamas).
The full results for all models are displayed in Table~\ref{tab:zero_shot_acc}.

As shown in Table~\ref{tab:acc_gap}, most LLMs exhibit negative $\text{Acc}_\text{gap}$ values, indicating that $\text{Acc}_\text{neg}$ consistently exceeds $\text{Acc}_\text{pos}$. 
This pattern reflects LLMs' inherent bias toward skepticism, showing a stronger inclination to reject rather than accept medical judgments, which aligns with previous research findings~\cite{sun-etal-2024-head, xiong2024can}. 
This behavior may be due to the model's intrinsic uncertainty in the domain-specific knowledge required to validate medical statements, which suggests a shortage of knowledge in medicine and healthcare. 

\subsection{RQ2: How well are LLMs calibrated in medical and healthcare contexts?}
\label{sec:rq2_calibration}

Given the bias in accuracies observed in Section~\ref{sec:rq1}, we aim to assess the calibration of LLMs~\cite{guo2017calibration, minderer2021revisiting, xiao2022uncertainty} in this section.

Miscalibration is commonly quantified in terms of Expected Calibration Error (ECE)~\cite{naeini2015obtaining}, which measures the absolute difference between predictive confidence and accuracy.
For better visualization, we present the calibration curve by plotting the pairs of confidence and accuracy.

To illustrate, considering \(N\) samples with predicted probabilities \(p_i\) and true labels \(y_i\) (\(y_i \in \{0,1\}\)), we plot the calibration curve as follows. 
Firstly, the interval \([0,1]\) is divided into \(M\) equal-width bins (we set $M=20$ in our experiment). Define the \(k\)-th bin \(B_k\) as 
\begin{equation*}
B_k = \{ i \mid p_i \in \left[\frac{k-1}{M}, \frac{k}{M}\right) \}, k = 1, 2, \dots, M.
\end{equation*}
Then for each bin \(B_k\), we compute the average predicted probability (confidence score) as 
\begin{equation*}
\hat{p}_k = \frac{1}{|B_k|} \sum_{i \in B_k} p_i,
\end{equation*}
and the observed positive rate (accuracy) as 
\begin{equation*}
\hat{y}_k = \frac{1}{|B_k|} \sum_{i \in B_k} y_i.
\end{equation*}
The calibration curve is then presented by plotting \(\{ (\hat{p}_k, \hat{y}_k) \}_{k=1}^{M}\). 
A perfectly calibrated model would have these points lie on the diagonal $y=x$.

\begin{figure}[h]
    \centering
    \begin{subfigure}{0.233\textwidth}
        \centering
        \includegraphics[width=\textwidth]{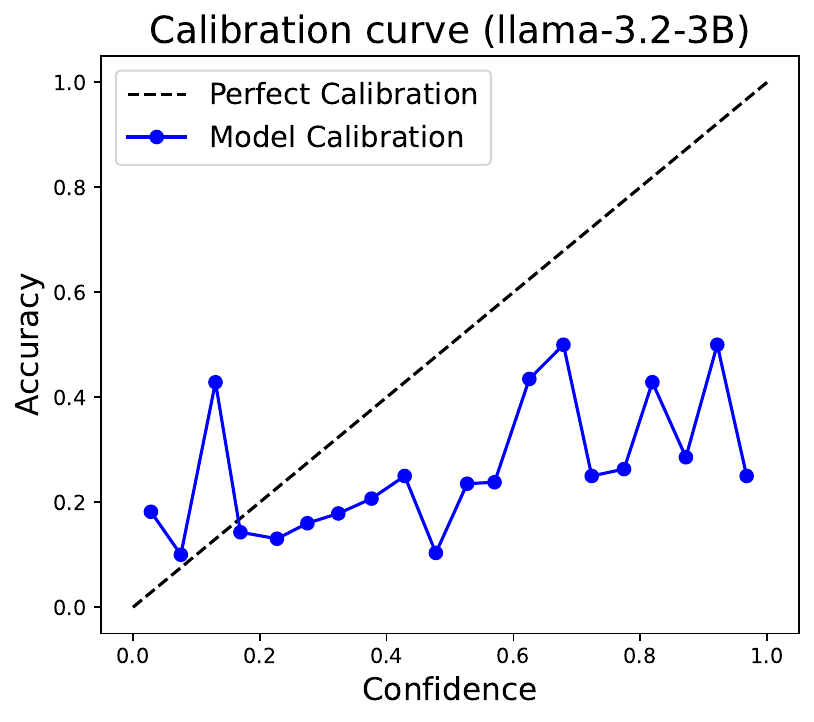}
        \caption{Calibration curve for \llamad.}
    \end{subfigure}
    \hfill
    \begin{subfigure}{0.233\textwidth}
        \centering
        \includegraphics[width=\textwidth]{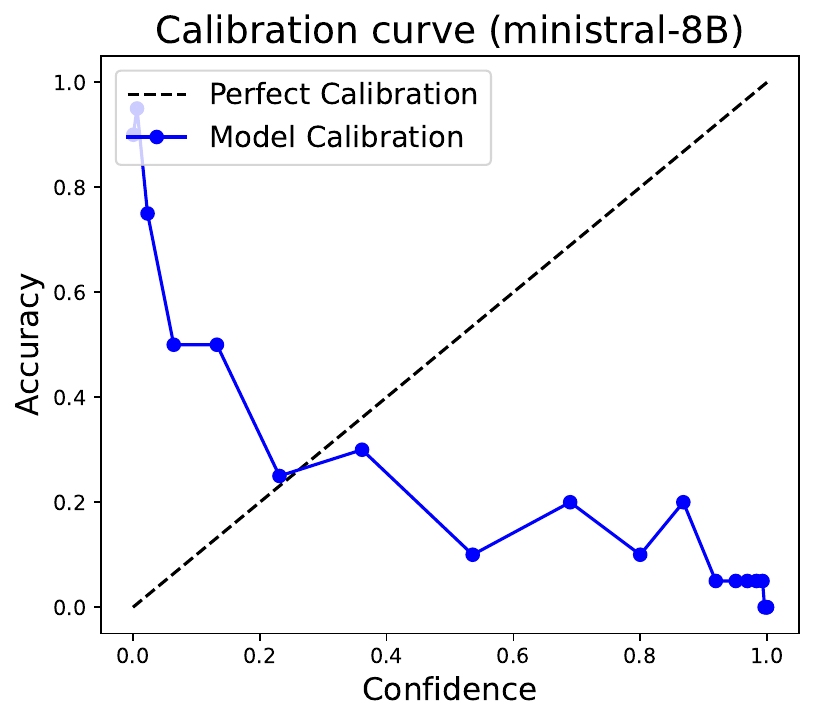}
        \caption{Calibration curve for \ministral.}
    \end{subfigure}
    
    \begin{subfigure}{0.233\textwidth}
        \centering
        \includegraphics[width=\textwidth]{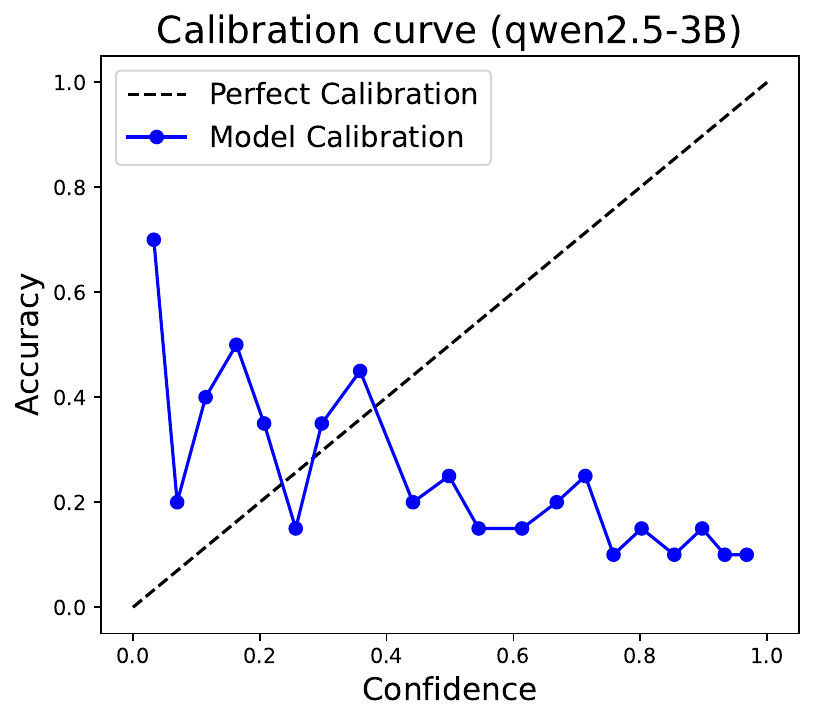}
        \caption{Calibration curve for \qwenc.}
    \end{subfigure}
    \hfill
    \begin{subfigure}{0.233\textwidth}
        \centering
        \includegraphics[width=\textwidth]{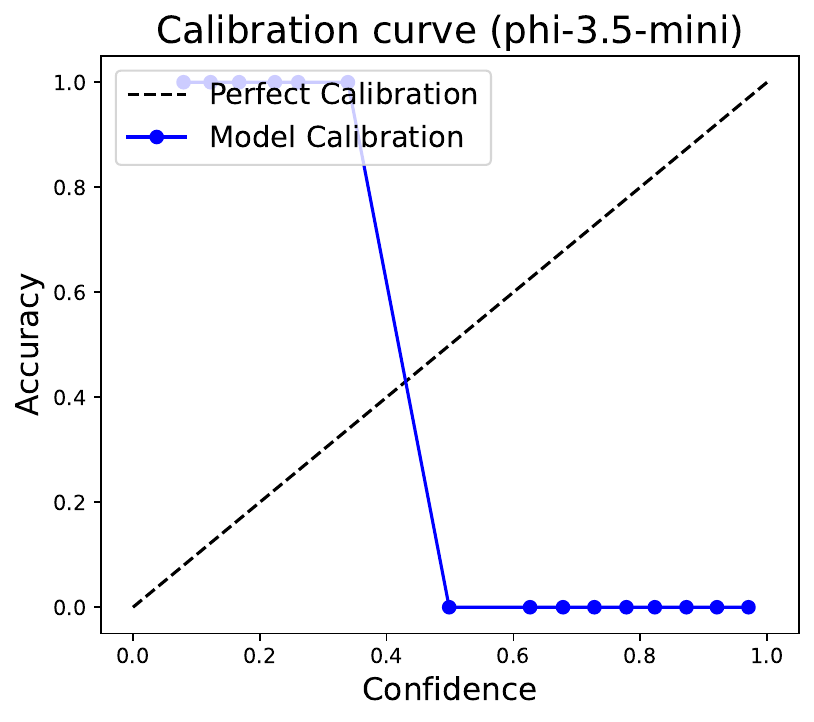}
        \caption{Calibration curve for \phic.}
    \end{subfigure}
    
    \begin{subfigure}{0.233\textwidth}
        \centering
        \includegraphics[width=\textwidth]{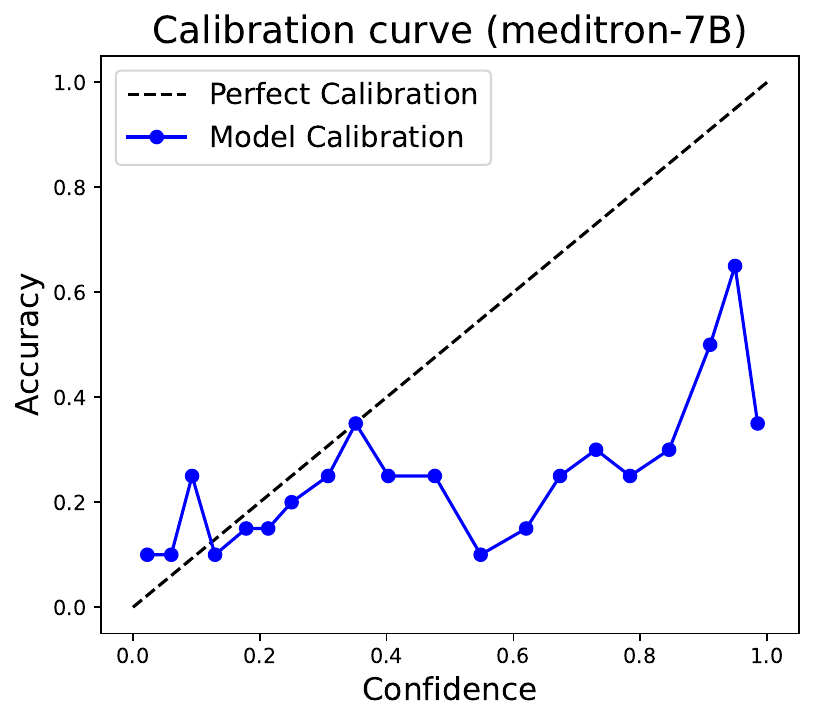}
        \caption{Calibration curve for \meditron.}
    \end{subfigure}
    \hfill
    \begin{subfigure}{0.233\textwidth}
        \centering
        \includegraphics[width=\textwidth]{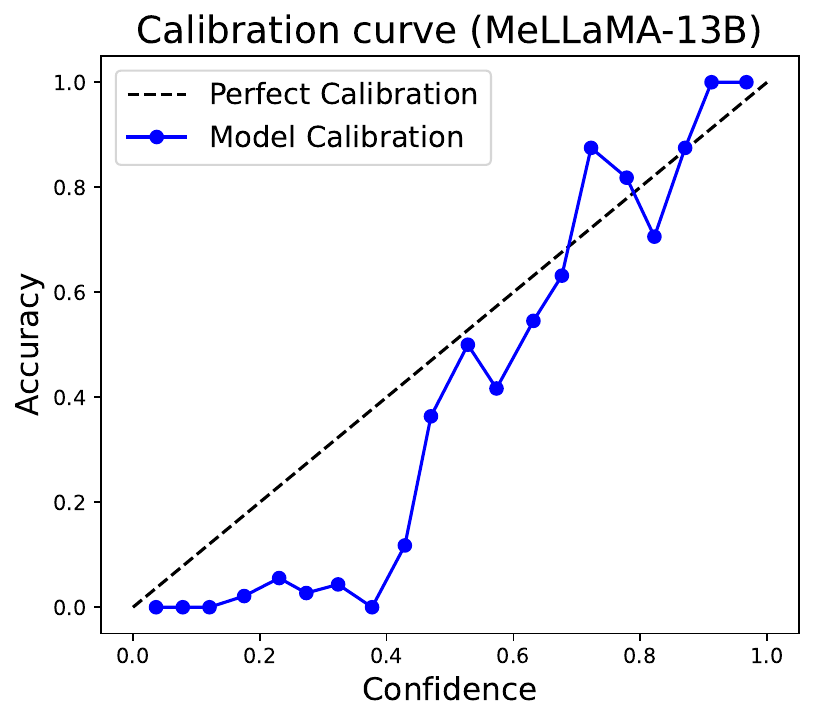}
        \caption{Calibration curve for \mellama.}
    \end{subfigure}
    
    \caption{Calibration curves for representative models from four different open-source LLM families (Llama, Mistral, Qwen, and Phi), as well as Medical LLMs (\meditron and \mellama).}
    \label{fig:calibration_curves}
    \vspace{-1em}
\end{figure}

The calibration curves of representative models from four general-domain LLM families and medical LLMs are shown in Figure~\ref{fig:calibration_curves}. 
The results for all LLMs can be found in Figure~\ref{fig:app_calibration_curve_1} and Figure~\ref{fig:app_calibration_curve_2} in the Appendix~\ref{app:calibration}.

The LLMs tested in this study illustrate varying degrees of calibration.
We can observe that general-domain LLMs exhibit significant disparities from the perfect calibration line, while medical LLMs demonstrate better calibration performances.

For general-domain LLMs, they reveal erratic status, exhibiting both overconfidence and underconfidence patterns. It is especially notable for \phic, which may be due to its training process as other models in \phis also observed similar patterns as shown in Figure ~\ref{fig:app_calibration_curve_2}.
For \llamad, although it occasionally approaches the ideal calibration line when confidence is low, its uncertainty level is still high when the confidence increases.

For medical LLMs, the calibration of \meditron is far from satisfactory, particularly in the mid-confidence range (0.4-0.8), where it tends to be overconfident. Its accuracy improves at higher confidence scores but still remains inconsistent. 
As for \mellama, the curve shows better calibration overall, with predictions closely aligned with the perfect calibration in mid-to-high confidence range although it still tends to be overconfident at low confidence levels. 

Overall, our analysis reveals pervasive calibration deficiencies across all evaluated models on the \mkj dataset. Poor calibration can result in overconfidence, which may lead to misplaced trust in incorrect predictions~\cite{xiong2024can, chen-etal-2023-close, li2025fact}, or underconfidence, which could cause clinicians to overlook valuable insights~\cite{yang2010nurses, jiang2012calibrating}. 
This points to calibration as a critical area for LLMs in healthcare and medicine.

\subsection{RQ3: What are the underlying reasons behind LLMs' failure to retain certain critical medical knowledge?}
\label{sec:rq3}

Given that LLMs exhibit problems observed in Section \ref{sec:rq1} and \ref{sec:rq2_calibration}, a natural question is raised, "What are the underlying reasons behind LLMs' failure to retain or recall certain medical knowledge?"

\begin{table}[H]
    \centering
    \resizebox{0.45\textwidth}{!}{
    \begin{tabular}{lc}
        \toprule
        \textbf{Semantic Type} &\textbf{Accuracy} \\ 
        \midrule
        \multicolumn{2}{c}{\textbf{Top-5 most error-prone}} \\
        Sign or Symptom (SS) & 0.41 \\ 
        Hormone (H) & 0.55 \\ 
        Physiologic Function (PF) & 0.61 \\ 
        Neoplastic Process (NP) & 0.64 \\ 
        Congenital Abnormality (CA) & 0.65  \\ 

        \noalign{\vskip 5pt}
        \cline{1-2}
        \noalign{\vskip 5pt}
        
        \multicolumn{2}{c}{\textbf{Top-5 least error-prone}} \\
        Amino Acid, Peptide, or Protein (APP) & 0.95 \\ 
        Immunologic Factor (IF) & 0.95 \\ 
        Temporal Concept (TC) & 0.95\\  
        Clinical Drug (CD) & 0.93 \\ 
        Diagnostic Procedure (DP) & 0.88 \\  
        \bottomrule
    \end{tabular}
    }
    \caption{Top-5 most and least error-prone semantic types for LLMs. We denote the semantic types using their abbreviations in parentheses.}
    \label{tab:top_5_case}
    \vspace{-0.5em}
\end{table}

To explore this, we analyze the performances of LLMs when tasked with assessing entities categorized under different semantic types, and summarize the accuracies accordingly in the Table~\ref{tab:top_5_case}, where we find the performance gap can be as large as 50\%, and identify that questions involving \textit{Sign or Symptom} and \textit{Hormone} categories exhibit notably higher error rates in the \mkj dataset.

\begin{figure}[h!]
    \centering
    \includegraphics[width=0.49\textwidth]{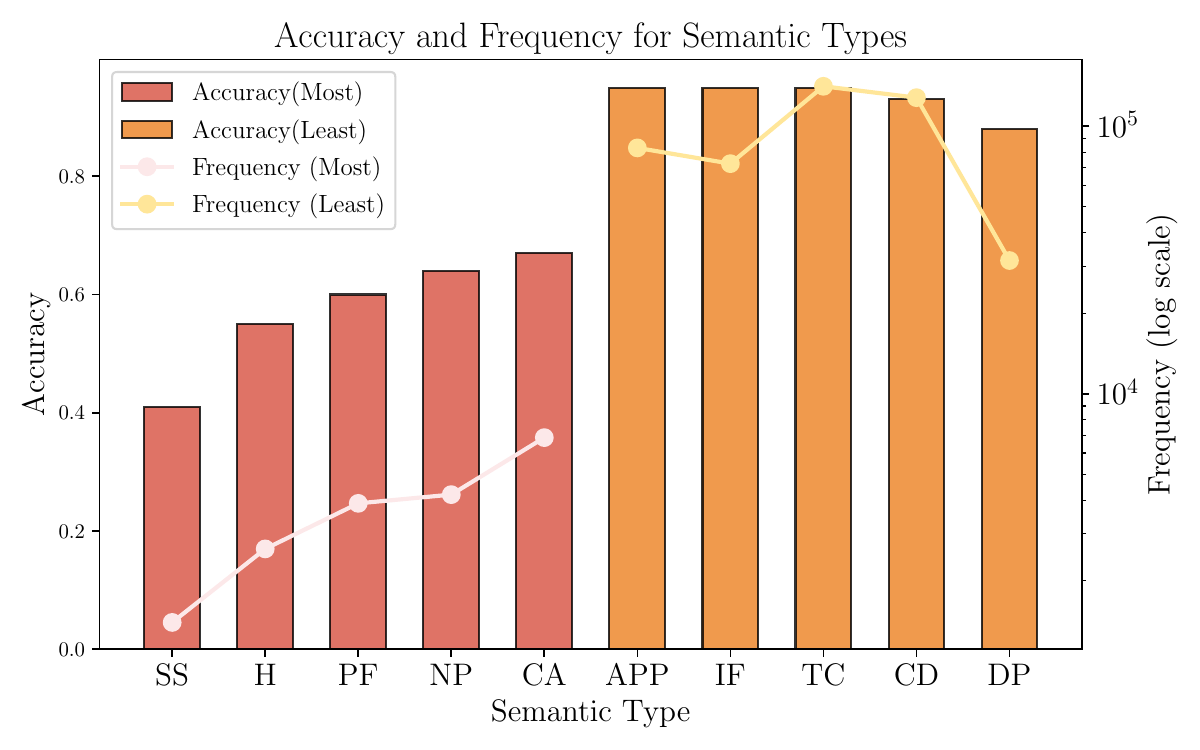}
    \caption{Overall accuracy (Acc) and frequency across most and least error-prone semantic types that are denoted via abbreviations as shown in Table~\ref{tab:top_5_case}.}
    \label{fig:acc_and_freq}
    \vspace{-1em}
\end{figure}

\begin{table*}[ht]
    \centering
    \vspace{-1.3em}
    \small
    \begin{tabular}{l|ccc|ccc}
        \toprule
        & \multicolumn{3}{c|}{\textbf{Sparse Retrieval}} & \multicolumn{3}{c}{\textbf{Dense Retrieval}} \\
        \textbf{Models} & \textbf{Acc} & \textbf{Acc$_\text{pos}$} & \textbf{Acc$_\text{neg}$} & \textbf{Acc} & \textbf{Acc$_\text{pos}$} & \textbf{$\text{Acc}_\text{neg}$} \\
        \midrule
        \gptthree & 
            90.75 \diff{90.75}{82.25}  & 
            89.00 \diff{89.00}{70.00}  & 
            91.33 \diff{91.33}{86.33}  & 
            88.75 \diff{88.75}{82.25}  & 
            87.00 \diff{87.00}{70.00}  & 
            89.33 \diff{89.33}{86.33} \\
        \gptfour & 
            93.50 \diff{93.50}{83.50}  & \textbf{99.00} \diff{99.00}{74.00} 
            & 91.66 \diff{91.66}{86.67}  & 92.50 \diff{92.50}{83.50} 
            & \textbf{99.00} \diff{99.00}{74.00}  & 90.33 \diff{90.33}{86.67} \\
        \gptfouro & 
            \textbf{96.50} \diff{96.50}{84.75}  & \textbf{99.00} \diff{99.00}{64.00} 
            & 95.66 \diff{95.66}{91.67}  & \textbf{95.50} \diff{95.50}{84.75} 
            & \textbf{99.00} \diff{99.00}{64.00}  & 94.33 \diff{94.33}{91.67} \\
        \midrule
        \llamaa & 
            88.75 \diff{88.75}{71.00}  & 
            69.00 \diff{69.00}{51.00}  & 
            95.33 \diff{95.33}{77.67}  & 
            87.00 \diff{87.00}{71.00}  & 
            64.00 \diff{64.00}{51.00}  & 
            94.66 \diff{94.66}{77.67} \\
        \llamab & 
            85.00 \diff{85.00}{68.25}  & 
            62.00 \diff{62.00}{51.00}  & 
            92.66 \diff{92.66}{74.00}  & 
            83.50 \diff{83.50}{68.25}  & 
            57.00 \diff{57.00}{51.00}  & 
            92.33 \diff{92.33}{74.00} \\
        \llamac & 
            69.25 \diff{69.25}{64.75}  & 
            \underline{20.00} \diff{20.00}{12.00}  & 
            85.66 \diff{85.66}{82.33}  & 
            67.75 \diff{67.75}{64.75}  & 
            \underline{29.00} \diff{29.00}{12.00}  & 
            80.66 \diff{80.66}{82.33} \\
        \llamad & 
            82.50 \diff{82.50}{66.75}  & 
            47.00 \diff{47.00}{42.00}  & 
            94.33 \diff{94.33}{75.00}  & 
            82.25 \diff{82.25}{66.75}  & 
            49.00 \diff{49.00}{42.00}  & 
            93.33 \diff{93.33}{75.00} \\
        \ministral & 
            90.00 \diff{90.00}{71.25}  & 
            67.00 \diff{67.00}{48.00}  & 
            97.66 \diff{97.66}{79.00}  & 
            91.50 \diff{91.50}{71.25}  & 
            73.00 \diff{73.00}{48.00}  & 
            97.66 \diff{97.66}{79.00} \\
        \qwena & 
            \underline{34.50} \diff{34.50}{56.00}  & 
            96.00 \diff{96.00}{50.00}  & 
            \underline{14.00} \diff{14.00}{58.00}  & 
            \underline{32.00} \diff{32.00}{56.00}  & 
            96.00 \diff{96.00}{50.00}  & 
            \underline{10.66} \diff{10.66}{58.00} \\
        \qwenb & 
            70.50 \diff{70.50}{53.00}  & 
            83.00 \diff{83.00}{73.00}  & 
            66.33 \diff{66.33}{46.33}  & 
            65.75 \diff{65.75}{53.00}  & 
            85.00 \diff{85.00}{73.00}  & 
            59.33 \diff{59.33}{46.33} \\
        \qwenc & 
            90.50 \diff{90.50}{68.25}  & 
            66.00 \diff{66.00}{16.00}  & 
            \textbf{98.66} \diff{98.66}{85.67}  & 
            91.50 \diff{91.50}{68.25}  & 
            71.00 \diff{71.00}{16.00}  & 
            \textbf{98.33} \diff{98.33}{85.67} \\
        \phia & 
            90.00 \diff{90.00}{69.50}  & 
            93.00 \diff{93.00}{64.00}  & 
            89.00 \diff{89.00}{71.33}  & 
            90.00 \diff{90.00}{69.50}  & 
            93.00 \diff{93.00}{64.00}  & 
            89.00 \diff{89.00}{71.33} \\
        \phib & 
            90.50 \diff{90.50}{71.75}  & 
            89.00 \diff{89.00}{58.00}  & 
            91.33 \diff{91.33}{76.33}  & 
            90.50 \diff{90.50}{71.75}  & 
            89.00 \diff{89.00}{58.00}  & 
            91.00 \diff{91.00}{76.33} \\
        \phic & 
            86.00 \diff{86.00}{69.25}  & 
            96.00 \diff{96.00}{67.00}  & 
            82.66 \diff{82.66}{70.00}  & 
            86.25 \diff{86.25}{69.25}  & 
            98.00 \diff{98.00}{67.00}  & 
            82.33 \diff{82.33}{70.00} \\
        \midrule
        \meditron & 
            91.75 \diff{91.75}{81.75}  & 
            84.00 \diff{84.00}{74.00}  & 
            94.33 \diff{94.33}{84.33}  & 
            88.50 \diff{88.50}{81.75}  & 
            81.00 \diff{81.00}{74.00}  & 
            91.00 \diff{91.00}{84.33} \\
        \mellama & 
            92.25 \diff{92.25}{83.25}  & 
            89.00 \diff{89.00}{79.00}  & 
            93.33 \diff{93.33}{84.67}  & 
            90.00 \diff{90.00}{83.25}  & 
            87.00 \diff{87.00}{79.00}  & 
            91.00 \diff{91.00}{84.67} \\
        \bottomrule
    \end{tabular}
    \caption{Performances of LLMs for RAG with sparse and dense retrieval methods. All numbers are in percentage (\%). Highest accuracies are in \textbf{bold}, and worst results are with \underline{underline}. The values in parentheses are the absolute differences between RAG and zero-shot, with $\uparrow$ denoting improvement and $\downarrow$ denoting decrement.}
    \label{tab:sparse_dense_rag}
    \vspace{-1em}
\end{table*}

We analyze that the observed performance degradation can be primarily attributed to two fundamental challenges in LLM training: long-tail knowledge distribution~\cite{pmlr-v202-kandpal23a} and co-occurrence bias~\cite{kang2023impact}. 
These issues manifest when medical concepts appear infrequently in training data, making it particularly challenging for LLMs to learn and retain accurate information about rare medical entities.

We validate this hypothesis through two complementary analyses. First, through comprehensive review of medical databases and literature~\cite{chen2024rarebench, wang2024assessing}, we find that the most error-prone cases consistently involve rare medical conditions. This is especially evident in semantic types like \textit{Sign or Symptom}, where uncommon terms such as "Swallow syncope" and "Asterixis" show high error rates, and \textit{Hormone}, where rare pharmaceutical compounds like "Prednisolone hexanoate" and "Fluprednisolone" prove particularly challenging for LLMs.

Second, to quantitatively validate the relationship between concept rarity and model performance, we conduct a systematic examination using the PubMed API\footnote{https://www.ncbi.nlm.nih.gov/books/NBK25497/}. 
By calculating the average frequency of medical terms per semantic type in the scientific literature, as illustrated in Figure~\ref{fig:acc_and_freq}, we reveal a clear correlation between term frequency and model accuracy. 
Semantic types with lower accuracies consistently demonstrate lower frequencies in the medical literature, providing empirical evidence for the impact of the long-tail knowledge.

These findings highlight a critical problem in current LLMs: their performances are significantly impacted by the uneven distribution of medical knowledge in training data, leading to substantial disparities in accuracy across different semantic types, particularly for rare medical concepts.

\subsection{RQ4: What strategies can enhance the response accuracy of LLMs?}
\label{sec:rq4}

Current results by zero-shot prompting show that there still exists a gap in response accuracy.
To explore more responsible approaches for utilizing LLMs, we experiment the technique of retrieval-augmented generation (RAG)~\cite{guu2020retrieval, karpukhin-etal-2020-dense, lewis2020retrieval}.

\paragraph{Experiment setup.} The experiment setup for RAG is outlined as below.
\begin{itemize}[left=0pt, topsep=0pt, itemsep=0pt, partopsep=0pt,parsep=0pt]
    \item Retriever. We incorporates both sparse retrieval method BM25~\cite{robertson2009probabilistic} and dense retrieval method Sentence-BERT~\cite{reimers-2019-sentence-bert, thakur-2020-AugSBERT}. 
    \item Documents. We utilize the constructed knowledge base $\mathcal{D}$ as described in Section~\ref{sec:data_collect} as our retrieval document corpus.
    \item Implementation. We retrieved documents for each query according to their cosine similarities, and prepend the top-5 to the input of the LLMs following~\citet{ram2023context} and \citet{shi-etal-2024-replug}. Detailed prompts can be found in Appendix~\ref{app:prompt}.
\end{itemize}

The experiment outputs are listed in Table~\ref{tab:sparse_dense_rag}. The performance differentials between RAG and zero-shot approaches are indicated in parentheses, computed as RAG results minus zero-shot results.


It is demonstrated that RAG enhances the response accuracy compared with zero-shot prompting on most cases, reaching 83.05\% accuracy with sparse retrieval and 81.85\% with dense retrieval on average, which greatly helps the application of LLMs in real-life settings.

However, we also observe that some small-size models such as \qwena and \llamac exhibit performance disparities between Acc$_\text{pos}$ and Acc$_\text{neg}$, which match the overconfidence patterns observed in Section~\ref{sec:rq2_calibration} and Figure~\ref{fig:app_calibration_curve_1}.
We hypothesize that this may be due to small models' inability to properly handle the knowledge between external documents and from internal parameters, which is known as knowledge conflicts~\cite{xu-etal-2024-knowledge-conflicts, su2024conflictbank}.

Our experiment underscores the importance of retrieval methods to complement the inherent strengths and weaknesses of LLMs~\cite{chen2024benchmarking}. 
By integrating effective RAG systems, we can significantly improve LLMs' ability in real-life medical QA scenarios.



\section{Conclusion}
We introduce the Medical Knowledge Judgment (\mkj) dataset which is designed to assess the factuality of LLMs in medical and healthcare by extracting and formulating knowledge triplets into one-hop direct judgment questions.
Our experiments on the \mkj dataset and subsequent analysis indicate that LLMs still face challenges in retaining factual knowledge, with notable variations in performances across different medical semantic types, especially for rare medical conditions.
In the meanwhile, most LLMs exhibit poor calibration status with mixed overconfidence and underconfidence patterns.
To address these issues, we investigate retrieval-augmented generation, and demonstrate its effectiveness in enhancing accuracy and reducing uncertainty in medical knowledge retention.



\section*{Limitations}
Despite the systematic construction of the \mkj dataset, evaluation experiment and subsequent analysis on LLMs, there are also limitations of our work.
\begin{itemize}
    \item Because of the large scale of UMLS, we begin with a set of existing multi-hop medical QA benchmarks. However, if the benchmarks adopted in our dataset construction pipeline do not comprehensively cover concepts or components in medicine and healthcare, the constructed dataset may also lack coverage on some type of entities to some extent.
    \item Due to the constraints on GPU devices, our devices cannot support run LLMs with very large sizes, which may limit our analysis and findings extend to larger models.
\end{itemize}


\bibliography{custom}

\clearpage
\appendix
\section{Appendix}
\label{sec:appendix}

\subsection{Dataset Details}
\label{app:data_detail}

\subsubsection{Basic information}
We extract triplets $(s, p, o)$ from UMLS, where there is a basic attribute called ``Semantic Type'' that denotes the type of an entity or concept. By doing substitution on the object of a triplet, we assess LLM's knowledge on the object $o$ with the predicate $p$ of subject $s$. Therefore, we use the semantic type of the object $o$ to denote the type of a judgment question. 

Specifically, there are 236 semantic types among the collected triplets $\mathcal{T}$. Due to the large amount of semantic types, we may only be able to list some of them here:
\textit{Organic Chemical}, 
\textit{Pharmacologic Substance}, 
\textit{Congenital Abnormality}, 
\textit{Finding}, 
\textit{Disease or Syndrome}, 
\textit{Virus}, 
\textit{Injury or Poisoning}, 
\textit{Sign or Symptom}, 
\textit{Therapeutic or Preventive Procedure}, 
\textit{Physiologic Function}, 
\textit{Idea or Concept}, 
\textit{Pathologic Function}, 
\textit{Functional Concept}, 
\textit{Molecular Function}, 
\textit{Amino Acid, Peptide, or Protein}, 
\textit{Immunologic Factor}, 
\textit{Neoplastic Process}, 
\textit{Professional or Occupational Group}, 
\textit{Clinical Drug}, 
\textit{Hormone}, 
\textit{Biologically Active Substance}, 
\textit{Indicator, Reagent, or Diagnostic Aid}, 
\textit{Intellectual Product}, 
\textit{Health Care Activity}, 
\textit{Body Part, Organ, or Organ Component}, 
\textit{Temporal Concept}, 
\textit{Diagnostic Procedure}, 
\textit{Body System}.

We further group our judgment questions in the \mkj dataset into three categories according to the semantic types, including \textit{Biomedical Entities}, \textit{Pathological Conditions}, and \textit{Clinical Practice}, as shown in Figure~\ref{fig:overview}.

\subsubsection{Statistics about \mkj dataset}
Here we provide some detailed statistics on our \mkj dataset. 

During the construction process, we collect 5820 high-quality triplets $\mathcal{T}$. There are more than 40,000 entities with 236 different semantic types. The average number of entities for each semantic type is around 160, which ensures our object substitution operation (Section~\ref{sec:data_sub}) in the data generation process is effective with little chance of introducing noise or mistake.

\begin{table*}[t!]
    \centering
    \resizebox{\textwidth}{!}{
        \begin{tabular}{l|c}
            \hline
            \textbf{Model} & \textbf{Version and Source} \\
            \toprule
            \gptthree & \url{https://platform.openai.com/docs/models/gpt-3-5-turbo\#gpt-3-5-turbo}\\
            \gptfour & \url{https://platform.openai.com/docs/models\#gpt-4o-mini} \\
            \gptfouro & \url{https://platform.openai.com/docs/models\#gpt-4o} \\
            \claudea & \url{https://docs.anthropic.com/en/docs/about-claude/models\#model-comparison-table} \\
            \claudeb & \url{https://docs.anthropic.com/en/docs/about-claude/models\#model-comparison-table} \\
            
            \midrule
            \llamaa & \url{https://huggingface.co/meta-llama/Meta-Llama-3-8B-Instruct} \\
            \llamab & \url{https://huggingface.co/meta-llama/Llama-3.1-8B-Instruct} \\
            \llamac & \url{https://huggingface.co/meta-llama/Llama-3.2-1B-Instruct}\\
            \llamad & \url{https://huggingface.co/meta-llama/Llama-3.2-3B-Instruct}\\
            \ministral & \url{https://huggingface.co/mistralai/Ministral-8B-Instruct-2410}\\
            \qwena & \url{https://huggingface.co/Qwen/Qwen2.5-0.5B-Instruct}\\
            \qwenb & \url{https://huggingface.co/Qwen/Qwen2.5-1.5B-Instruct}\\
            \qwenc & \url{https://huggingface.co/Qwen/Qwen2.5-3B-Instruct}\\
            \phia & \url{https://huggingface.co/microsoft/Phi-3-mini-4k-instruct}\\
            \phib & \url{https://huggingface.co/microsoft/Phi-3-mini-128k-instruct} \\
            \phic & \url{https://huggingface.co/microsoft/Phi-3.5-mini-instruct} \\

            \midrule
            \meditron & \url{https://huggingface.co/epfl-llm/meditron-7b} \\
            \mellama & \url{https://physionet.org/content/me-llama/1.0.0/MeLLaMA-13B/} \\
            \bottomrule
        \end{tabular}
    }
    \caption{Detailed versions and sources of LLMs used in our experiments.}
    \label{tab:app_LLMs}
\end{table*}

\subsection{Model Details}
\label{app:model_details}
\paragraph{Details of LLMs in our experiment.}
Some full names of LLMs may occupy too-much space in the paper main body. Therefore, we provide the detailed versions and sources of LLMs evaluated in our experiments, as listed in Table~\ref{tab:app_LLMs}.

\subsection{Prompts}
\label{app:prompt}

\paragraph{Prompt for Extracting Keywords.}
In section~\ref{sec:data_collect}, we extracting keywords in a problem for medical and healthcare, and the prompt is given below in Figure~\ref{fig:med_ner_prompt}.

\begin{figure}[ht]
    \centering
    \includegraphics[width=0.48\textwidth]{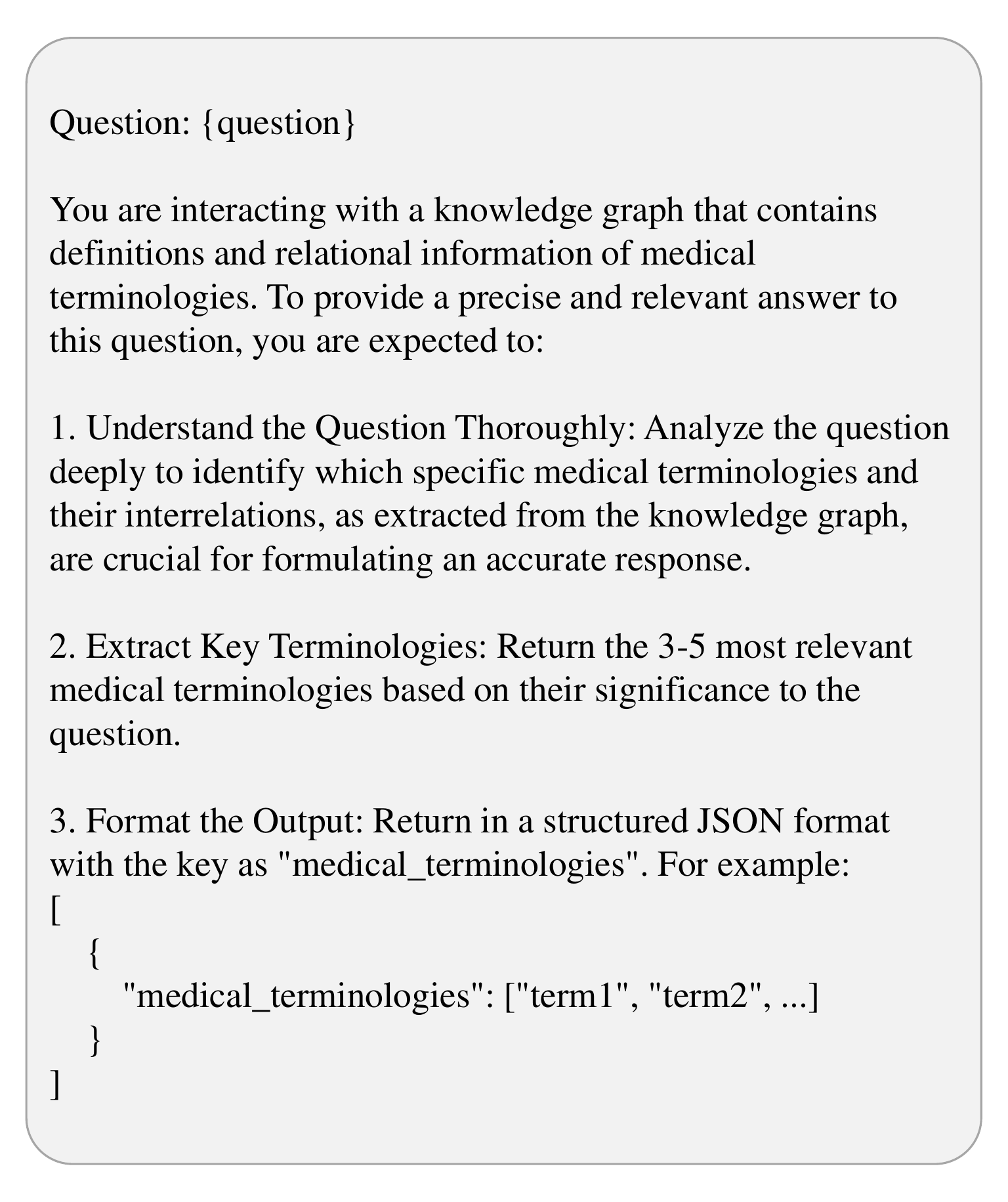}
    \caption{Prompt for extracting medical keywords.}
    \label{fig:med_ner_prompt}
\end{figure}

\paragraph{Prompt for Constructing Templates.}
In section~\ref{sec:data_temp}, we prompt LLMs to generate templates for constructing judgment questions. By utilizing few-shot prompting technique (we provide 3 examples in the context), LLMs can generate high-quality templates that are logical and coherent. The prompt is given below in Figure~\ref{fig:template_gen_prompt}.

\begin{figure}[!ht]
    \centering
    \includegraphics[width=0.48\textwidth]{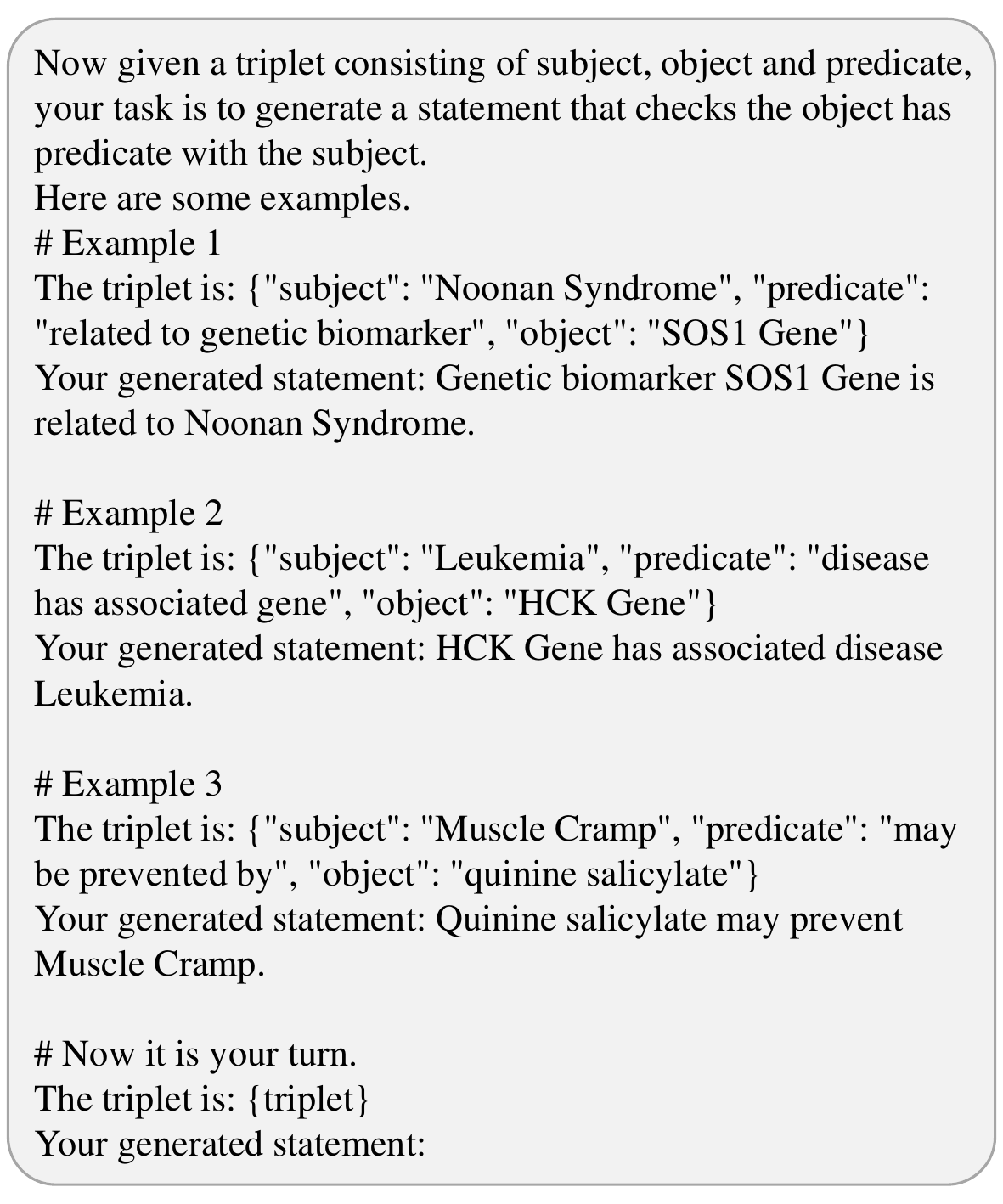}
    \caption{Prompt used for statement generation.}
    \label{fig:template_gen_prompt}
\end{figure}

\begin{figure}[!ht]
    \centering
    \includegraphics[width=0.48\textwidth]{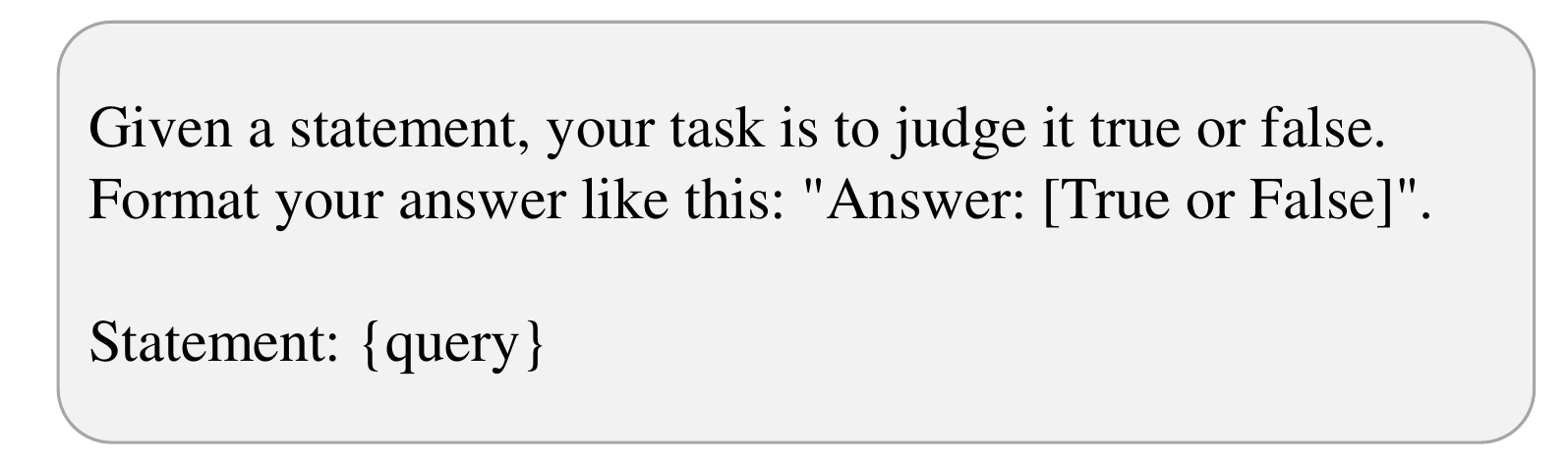}
    \caption{Zero-shot Prompting}
    \label{fig:zo_prompt}
\end{figure}

\begin{figure}[!h]
    \vspace{-1em}
    \centering
    \includegraphics[width=0.48\textwidth]{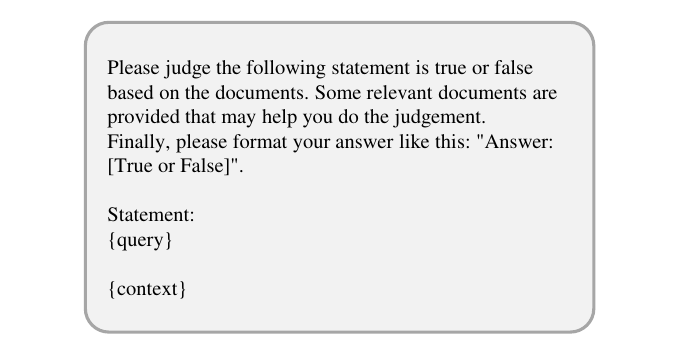}
    \caption{Prompt for retrieval-augmented generation}
    \label{fig:rag_prompt}
\end{figure}

\paragraph{Prompts for Evaluation.}
The prompts used for evaluation with zero-shot prompting and retrieval-augmented generation (RAG) are provided in Figure~\ref{fig:zo_prompt} and Figure~\ref{fig:rag_prompt}.

\subsection{Performances with zero-shot prompting}
\label{app:zero_shot_perf}
We display the full results of LLMs with zero-shot prompting in Table~\ref{tab:zero_shot_acc} below.

\begin{table}[h]
    \vspace{-1.6em}
    \centering
    
    \resizebox{0.45\textwidth}{!}{
    \begin{tabular}{lcccc}
      \toprule
      \textbf{Models} & \textbf{Acc} & \textbf{Acc$_{\text{pos}}$} & \textbf{Acc$_{\text{neg}}$} & \textbf{Acc$_{\text{gap}}$} \\
      \midrule
      \gptthree    & 0.82 & 0.70 & 0.86 & -0.16 \\
      \gptfour     & 0.84 & 0.74 & 0.87 & -0.13 \\
      \gptfouro    & 0.85 & 0.64 & 0.92 & -0.28 \\
      \claudea     & 0.72 & 0.88 & 0.66 &  0.22 \\
      \claudeb     & 0.83 & 0.68 & 0.88 & -0.20 \\
      \llamaa      & 0.71 & 0.51 & 0.78 & -0.27 \\
      \llamab      & 0.68 & 0.51 & 0.74 & -0.23 \\
      \llamac      & 0.65 & 0.12 & 0.82 & -0.70 \\
      \llamad      & 0.67 & 0.42 & 0.75 & -0.33 \\
      \ministral   & 0.71 & 0.48 & 0.79 & -0.31 \\
      \qwena       & 0.56 & 0.50 & 0.58 & -0.08 \\
      \qwenb       & 0.53 & 0.73 & 0.46 &  0.27 \\
      \qwenc       & 0.68 & 0.16 & 0.86 & -0.70 \\
      \phia        & 0.70 & 0.64 & 0.72 & -0.08 \\
      \phib        & 0.72 & 0.58 & 0.76 & -0.18 \\
      \phic        & 0.69 & 0.67 & 0.71 & -0.04 \\
      \meditron    & 0.82 & 0.74 & 0.84 & -0.10 \\
      \mellama     & 0.83 & 0.79 & 0.85 & -0.06 \\
      \bottomrule
    \end{tabular}
    }

    \caption{Model performances with zero-shot prompting.}
    \label{tab:zero_shot_acc}
\end{table}



\subsection{Calibration Curves and Uncertainty Quantification}
\label{app:calibration}
\begin{figure}[ht]
    \centering
    
    \begin{subfigure}{0.233\textwidth}
        \centering
        \includegraphics[width=\textwidth]{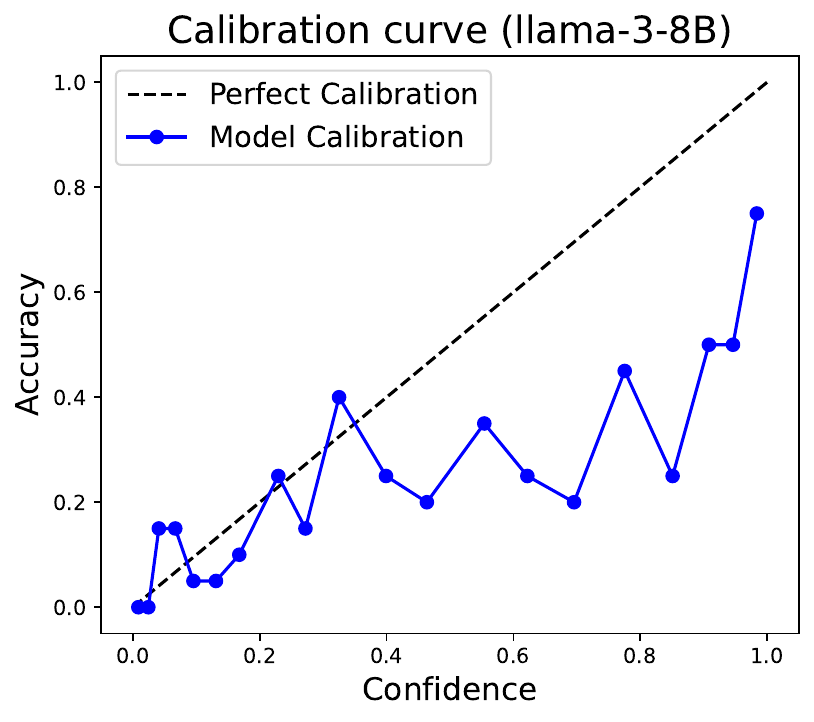}
        \caption{Calibration curve for \llamaa.}
        \label{fig:app_cal_fig1}
    \end{subfigure}
    \hfill
    \begin{subfigure}{0.233\textwidth}
        \centering
        \includegraphics[width=\textwidth]{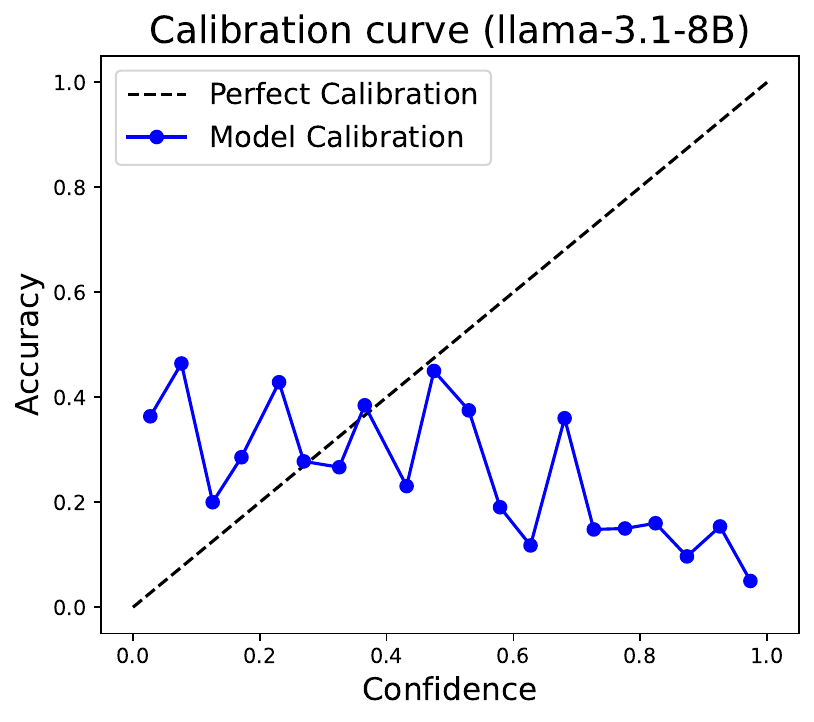}
        \caption{Calibration curve for \llamab.}
        \label{fig:app_cal_fig2}
    \end{subfigure}
    
    \vspace{0.5cm}
    
    \begin{subfigure}{0.233\textwidth}
        \centering
        \includegraphics[width=\textwidth]{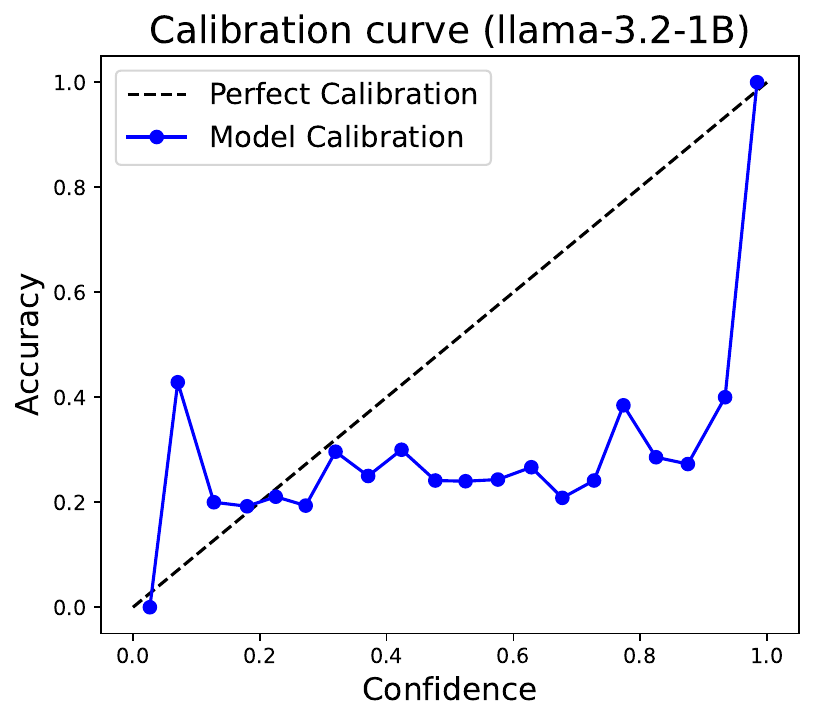}
        \caption{Calibration curve for \llamac.}
        \label{fig:app_cal_fig3}
    \end{subfigure}
    \hfill
    \begin{subfigure}{0.233\textwidth}
        \centering
        \includegraphics[width=\textwidth]{figures/calibration/calibration_curve_llama-3.2-3B.pdf}
        \caption{Calibration curve for \llamad.}
        \label{fig:app_cal_fig4}
    \end{subfigure}

    \vspace{0.5cm}

    \begin{subfigure}{0.233\textwidth}
        \centering
        \includegraphics[width=\textwidth]{figures/calibration/calibration_curve_ministral-8B.pdf}
        \caption{Calibration curve for \ministral.}
        \label{fig:app_cal_fig5}
    \end{subfigure}
    \hfill
    \begin{subfigure}{0.233\textwidth}
        \centering
        \includegraphics[width=\textwidth]{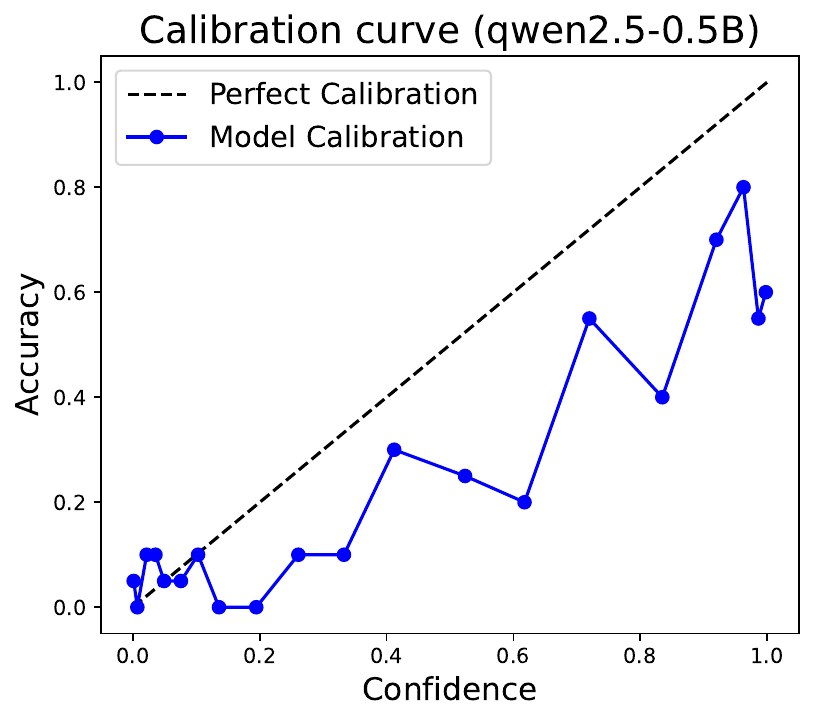}
        \caption{Calibration curve for \qwena.}
        \label{fig:app_cal_fig6}
    \end{subfigure}

    \caption{Calibration curves for all tested LLMs (first subfigure).}
    \label{fig:app_calibration_curve_1}
\end{figure}

\begin{figure}[htbp]
    \centering

    \begin{subfigure}{0.233\textwidth}
        \centering
        \includegraphics[width=\textwidth]{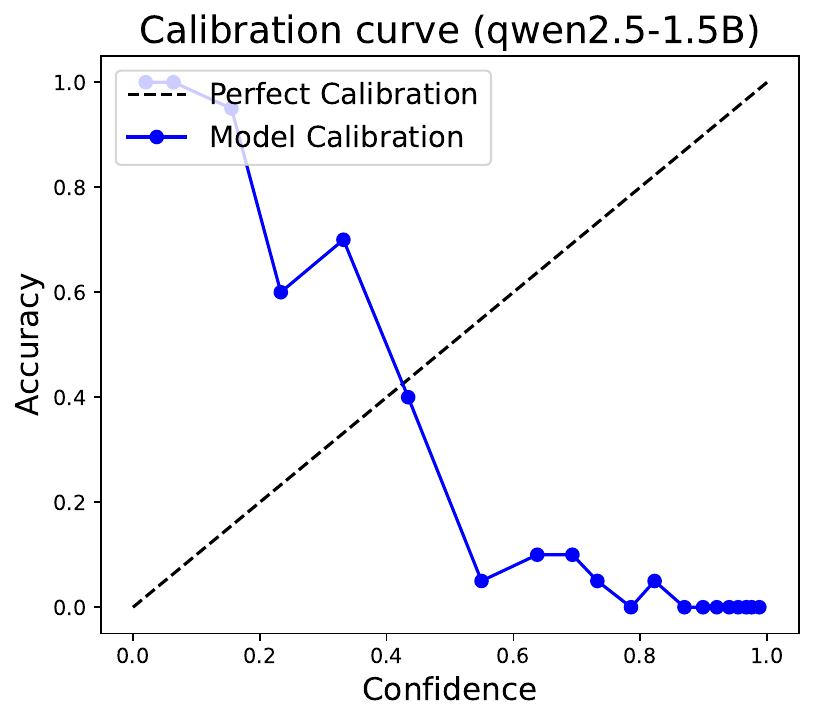}
        \caption{Calibration curve for \qwenb.}
        \label{fig:app_cal_fig7}
    \end{subfigure}
    \hfill
    \begin{subfigure}{0.233\textwidth}
        \centering
        \includegraphics[width=\textwidth]{figures/calibration/calibration_curve_qwen2.5-3B.pdf}
        \caption{Calibration curve for \qwenc.}
        \label{fig:app_cal_fig8}
    \end{subfigure}

    \vspace{0.5cm}
    
    \begin{subfigure}{0.233\textwidth}
        \centering
        \includegraphics[width=\textwidth]{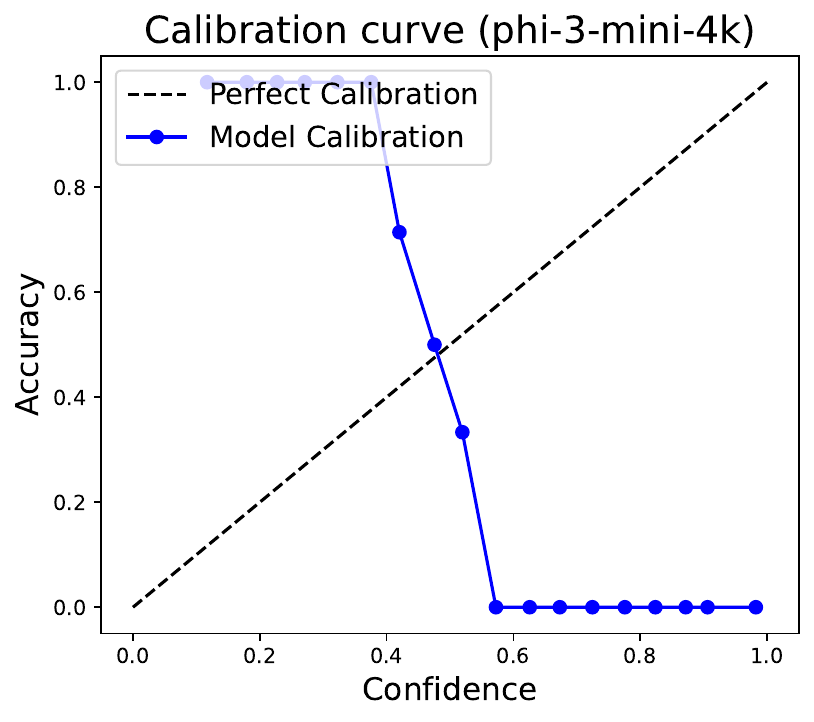}
        \caption{Calibration curve for \phia.}
        \label{fig:app_cal_fig9}
    \end{subfigure}
    \hfill
    \begin{subfigure}{0.233\textwidth}
        \centering
        \includegraphics[width=\textwidth]{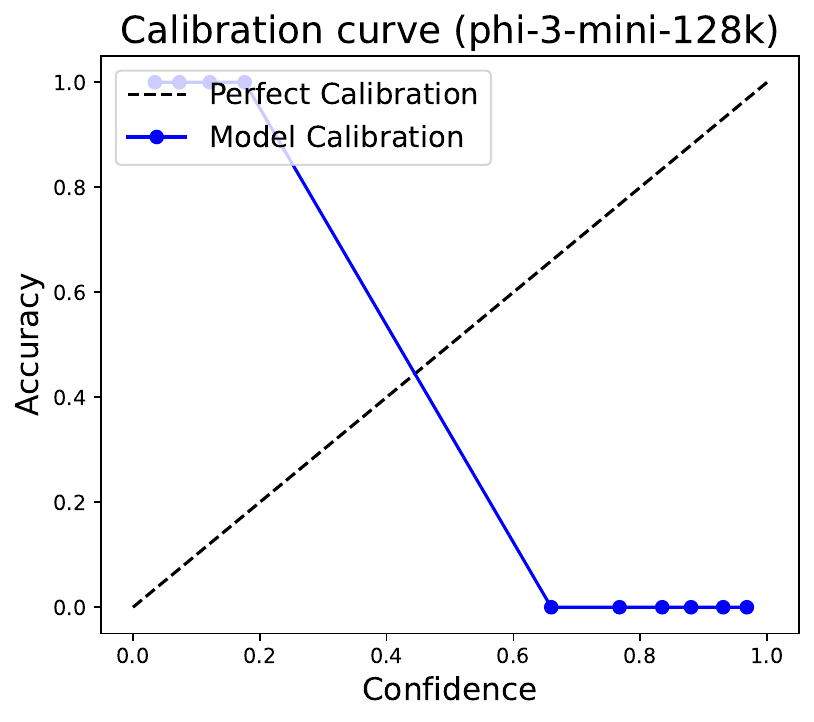}
        \caption{Calibration curve for \phib.}
        \label{fig:app_cal_fig10}
    \end{subfigure}

    \vspace{0.5cm}

    \begin{subfigure}{0.233\textwidth}
        \centering
        \includegraphics[width=\textwidth]{figures/calibration/calibration_curve_phi-3.5-mini.pdf}
        \caption{Calibration curve for \phic.}
        \label{fig:app_cal_fig11}
    \end{subfigure}
    \hfill
    \begin{subfigure}{0.233\textwidth}
        \centering
        \includegraphics[width=\textwidth]{figures/calibration/calibration_curve_meditron-7B.pdf}
        \caption{Calibration curve for \meditron.}
        \label{fig:app_cal_fig12}
    \end{subfigure}
    
    \vspace{0.5cm}

    \begin{subfigure}{0.233\textwidth}
        \centering
        \includegraphics[width=\textwidth]{figures/calibration/calibration_curve_MeLLaMA-13B.pdf}
        \caption{Calibration curve for \mellama.}
        \label{fig:app_cal_fig13}
    \end{subfigure}

    \caption{Calibration curves for all tested LLMs (second subfigure).}
    \label{fig:app_calibration_curve_2}
\end{figure}

We list the calibration curves of all open-source LLMs we tested in this paper in Figure~\ref{fig:app_calibration_curve_1} and Figure~\ref{fig:app_calibration_curve_2}, including LLM families of Llama, Mistral, Qwen, Phi, and meidcal LLMs Meditron-7B and MeLLaMA-13B.

\end{document}